\documentclass[10pt,twocolumn,letterpaper]{article}
\pdfoutput=1

\usepackage[pagenumbers]{cvpr} 
\usepackage{graphicx}
\usepackage{booktabs}
\usepackage{array}
\usepackage{makecell}
\usepackage[accsupp]{axessibility}  

\definecolor{rankred}{HTML}{FED5D5}
\definecolor{rankblue}{HTML}{CCE2FF}

\definecolor{cvprblue}{rgb}{0.21,0.49,0.74}
\usepackage[pagebackref,breaklinks,colorlinks,allcolors=cvprblue]{hyperref}
\usepackage{multirow}
\newcommand{\bench}{OVBench}
\newcommand{\benchname}{OVBench}
\newcommand{\model}{VideoChat-Online}

\usepackage{colortbl}
\usepackage{marvosym}
\newcommand{\FPava}{AA}	
\newcommand{\Halava}{AP}	
\newcommand{\PMava}{AR}	
\newcommand{\SPava}{AL}	
\newcommand{\STPava}{AT}	
\newcommand{\TPava}{AS}	
\newcommand{\FPevent}{GSP}	
\newcommand{\Halevent}{SV}	
\newcommand{\PMevent}{PR}	
\newcommand{\TPevent}{SL}	
\newcommand{\FPtao}{MP}	
\newcommand{\Haltao}{OP}	
\newcommand{\PMtao}{TR}	
\newcommand{\SPtao}{OP}	
\newcommand{\STPtao}{OT}	
\newcommand{\TPtao}{OES}
\usepackage{xcolor}
\usepackage{tabularx}

\title{Online Video Understanding: OVBench and VideoChat-Online}

\author{
    Zhenpeng Huang$^{1\dagger}$, 
    Xinhao Li$^{1,3\dagger}$,
    Jiaqi Li$^{2\dagger}$,
    Jing Wang$^{1}$, 
    Xiangyu Zeng$^{1,3}$ \\
     Cheng Liang$^{1}$, Tao Wu$^{1}$, Xi Chen$^{2}$, Liang Li$^{2}$, Limin Wang$^{1,3,}$\textsuperscript{\Letter} \\
   ~~~
    \small$^1$State Key Laboratory for Novel Software Technology, Nanjing University~~~\\
    \small$^2$China Mobile Research Institute~~~
    \small$^3$OpenGVLab, Shanghai AI Laboratory \\
    {\small \url{https://videochat-online.github.io/}}
}

\begin{document}
\maketitle
\begin{abstract}

Multimodal Large Language Models (MLLMs) have significantly progressed in offline video understanding. However, applying these models to real-world scenarios, such as autonomous driving and human-computer interaction, presents unique challenges due to the need for real-time processing of continuous online video streams. To this end, this paper presents systematic efforts from three perspectives: evaluation benchmark, model architecture, and training strategy. First, we introduce \textbf{\bench{}}, a comprehensive question-answering benchmark designed to evaluate models' ability to perceive, memorize, and reason within online video contexts. It features 6 core task types across three temporal contexts—past, current, and future—forming 16 subtasks from diverse datasets. Second, we propose a new \textbf{P}yramid \textbf{M}emory \textbf{B}ank (\textbf{PMB}) that effectively retains key spatiotemporal information in video streams. Third, we proposed an offline-to-online learning paradigm, designing an interleaved dialogue format for online video data and constructing an instruction-tuning dataset tailored for online video training. This framework led to the development of \textbf{VideoChat-Online}, a robust and efficient model for online video understanding. Despite the lower computational cost and higher efficiency, VideoChat-Online outperforms existing state-of-the-art offline and online models across popular offline video benchmarks and \bench, demonstrating the effectiveness of our model architecture and training strategy. 
\end{abstract}

{
\renewcommand{\thefootnote}%
{\fnsymbol{footnote}}
\footnotetext[0]{$\dagger$ Equal contribution. \Letter{} Corresponding author.} 
}
\section{Introduction}

With the rapid development of Multimodal Large Language Models (MLLMs)~\cite{gpt4,gpt4o,videochat,videollama,pllava} in recent years, these models have demonstrated impressive performance on video understanding benchmarks~\cite{mvbench,egoschema,perception_test}. These advancements have laid the foundation for exploring real-time, online video scenarios, including autonomous driving, robotic assistants, and surveillance systems. Recent research, such as GPT-4o~\cite{gpt4o} and VideoLLM-Online~\cite{videollm_online}, Flash-VStream~\cite{flashvstream}, have further investigated online video understanding and model efficiency in streaming scenarios, highlighting the potential of MLLMs in understanding online video streams.
\begin{figure}
    \centering
    \includegraphics[width=0.5\textwidth]{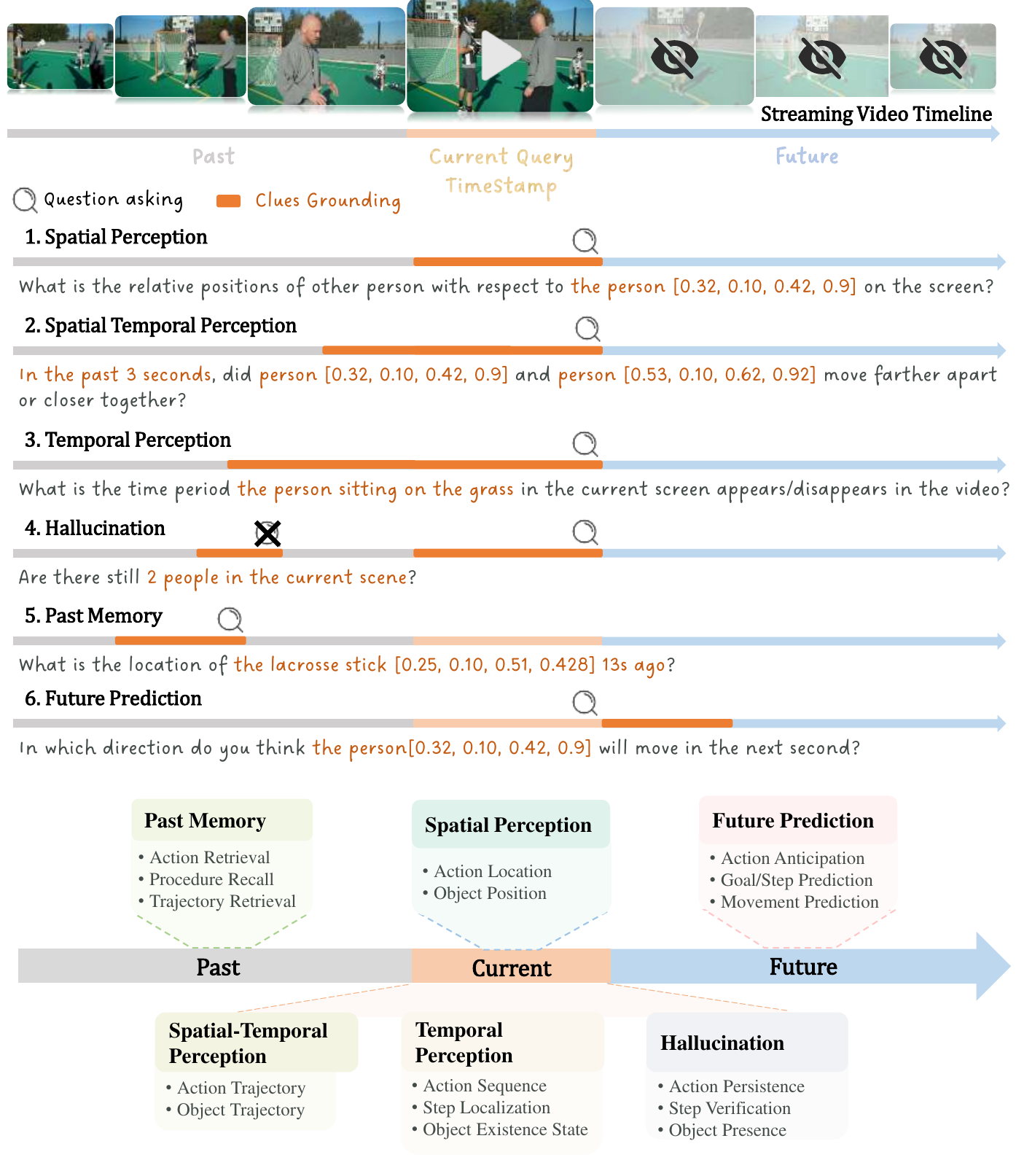}
    \caption{\bench~contains 6 core spatiotemporal understanding tasks in online scenarios, incorporating three primary temporal contexts—past, current, and future. Based on various interaction types, it is expanded into 16 subtasks in total.}
    \label{fig:tasks}
\end{figure}

Despite these advances, applying MLLMs to real-world streaming scenarios presents unique challenges. Offline processing refers to models that analyze entire videos post-capture, rather than responding in real-time as frames are received. The unique characteristics of online video streams are not fully considered by the existing works as follows:
\begin{itemize}
    \item \textbf{Online Temporal Perspective}: Based on the time when a user poses a question, the temporal perspective of online video streams can be distinctly defined as past, current, and future. In contrast to offline videos, it enables a finer temporal perspective (e.g., a few seconds ago, right now).
    \item \textbf{Time-dependent Contexts}: In offline video understanding, answers are derived from all prior frames, typically yielding a unique response. In streaming scenarios, as the temporal context evolves, the answer dynamically changes. Questions like "What is the person doing now?" may receive varying responses over time.
    \item \textbf{Real-time Spatio-temporal Interaction}: Applications like augmented reality (AR) glasses and autonomous driving systems require precise, real-time spatiotemporal interaction with the environment, where immediate responses to the environment (e.g. actions, objects, and events) are essential for functionality and safety.
    \item \textbf{Processing of infinitely long visual information}: Online video streams continuously introduce an infinite influx of new visual information. Therefore, designing online model architectures that can process and retain key information, akin to human cognition, is critical.
    
\end{itemize}

Given that most of the current video understanding benchmarks~\cite{mvbench, videomme, tgifqa, activitynet_qa, msrvtt_qa, MoVQA23, zhou2024mlvu, wang2024lvbench} are conducted in offline mode, there is a pressing need to build a benchmark specifically tailored to online video streams, taking into account their unique spatiotemporal characteristics. To tackle these problems, we introduce the \textbf{O}nline \textbf{V}ideo Understanding \textbf{Bench}mark, \textbf{\bench}. This benchmark aims to evaluate a model’s capacity to understand and interpret spatiotemporal details in online scenarios. As shown in \cref{fig:tasks}, we define the temporal context for streaming videos, and based on these temporal contexts, we design 6 task types encompassing a total of 16 subtasks. These tasks are based on seven datasets spanning 6 different domains (Movie, Instructional, Road, Scenes (Outdoor \& Indoor), and Open-domain) to ensure a diverse range of task scenarios. To create high-quality annotations, our benchmark employs human annotators who generate $\sim$7,000 high-quality annotations that emphasize spatiotemporal details. 

We evaluate leading MLLMs on \bench{}, including offline image/video MLLMs (adapted to streaming via sliding windows) and online video MLLMs. Current models exhibit poor online spatiotemporal understanding, with online models lagging significantly behind MLLMs. This motivates us to develop a strong baseline for online video understanding with a novel architecture and training strategy.
\textbf{(1) New Model Architecture}: Existing architectures~\cite{moviechat,moviechat+,videollm_online,flashvstream} struggle with fine-grained spatial details and long-range temporal dependencies as streaming contexts grow. We propose a Pyramid Memory Bank (PMB) to balance spatial and temporal understanding via progressive abstraction. PMB preserves recent-frame details while efficiently abstracting distant-frame information using adaptive frame eviction and resolution scaling, optimizing both comprehension and memory efficiency.
\textbf{(2) New Training Strategy}: A key limitation of existing MLLMs is the lack of a tailored training strategy. We introduce an offline-to-online learning paradigm, constructing interleaved dialogue-style online video instruction tuning data. This, combined with offline video data, progressively enhances both offline and online video understanding. 

Based on the aforementioned design principles, we have developed an efficient 4B-parameter online video MLLM, coined as \textbf{\model}, specifically designed for mobile deployment, thereby broadening the potential applications of online video understanding. On \bench, our model outperforms the open-source offline MLLM Qwen2-VL \cite{wang2024qwen2vl} (7B parameters) by \textbf{4.19\%} and the online Video MLLM Flash-Vstream~\cite{flashvstream} by \textbf{23.7\%}, achieving this with a more efficient architecture. Furthermore, it demonstrates state-of-the-art performance on established offline video benchmarks, highlighting its robustness across both online and offline video understanding. All the models and data are publicly available. We hope that this work's benchmark, dataset, and model will inspire future research on online video understanding. 

\section{Related Work}
\begin{figure*}[ht!]
    \centering
    \includegraphics[width=0.90\textwidth]{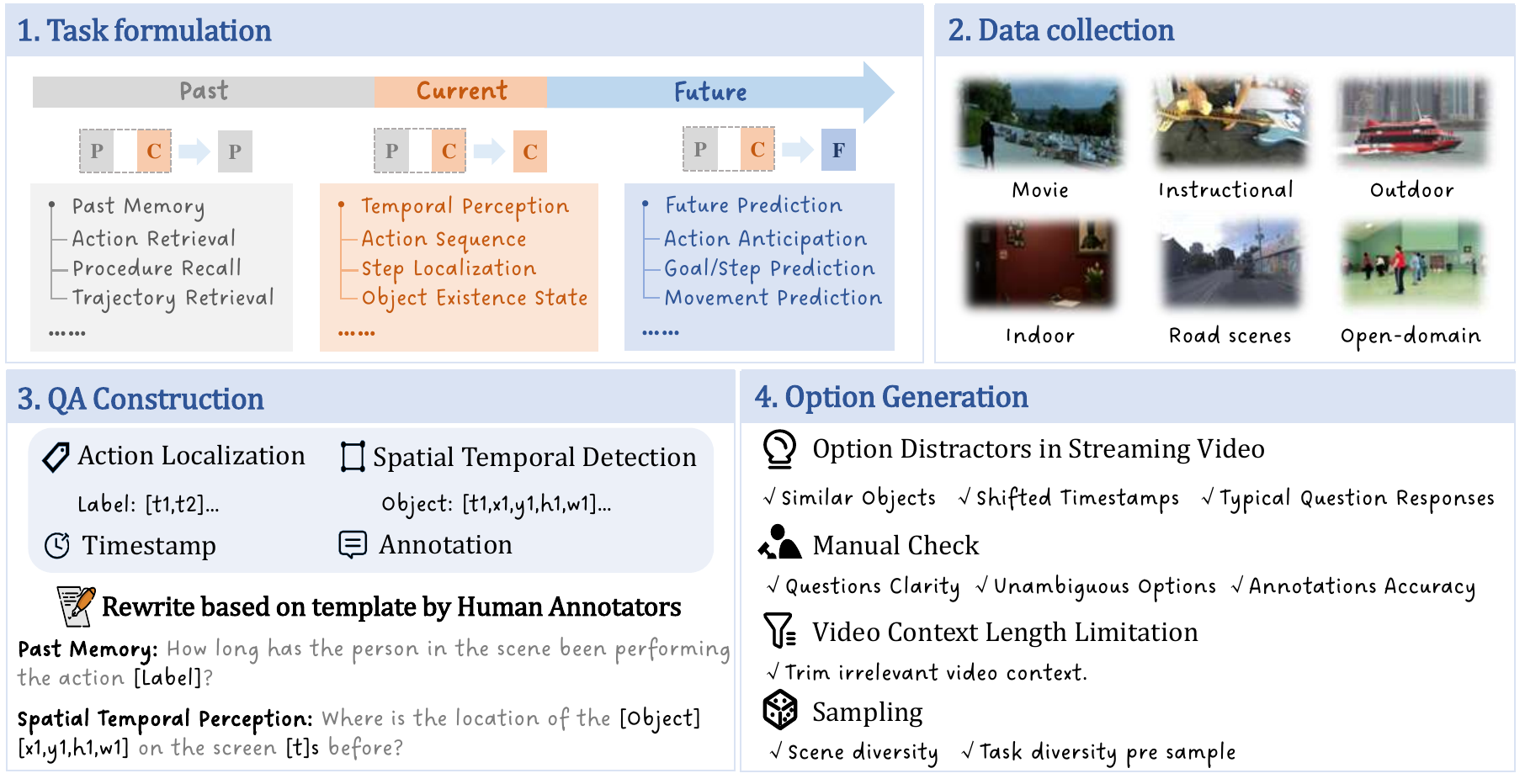}
    \caption{
    \textbf{Generation pipeline of \bench.} 
    We developed a method to ensure the quality of annotation based on the existing high-quality spatiotemporal data, including task definition, data collection, QA construction, and multiple-choice question generation suitable for streaming video scenarios. The details will be discussed in Section \ref{sec:ovbench}.
    }
    \label{fig:pipeline}
    \vspace{-0.3cm}
\end{figure*}
\noindent\textbf{Online Video MLLMs.} 
The advent of large language models (LLMs)~\cite{chatgpt, llama2, llama3, mistral, vicuna} has spurred substantial progress in multimodal understanding. Recent multimodal LLMs~\cite{wang2024qwen2vl, hu2024minicpm, timechat, LLaVA-OneVision, huang2024vtimellm, chen2023internvl, video_chatgpt, videochat, videochat2} have exhibited impressive capabilities in offline video comprehension by integrating visual encoders with LLMs. However, these models are inherently challenged in real-time applications due to their limited capacity for efficient streaming video frame compression, leading to increased computational demands and latency with accumulating input frames. Several strategies~\cite{moviechat, moviechat+,llamavid,chatunivi} have been explored to mitigate computational burden through video redundancy reduction. However, most of these models lack a design specifically tailored for online video stream processing. Recent studies have introduced MLLMs specifically designed for online stream understanding. VideoLLM-Online~\cite{videollm_online} pioneers the development of general-purpose AI assistants for real-time video stream dialogue and multi-task execution. However, its performance is limited by restricted per-frame visual token input due to the lack of effective streaming context compression. VideoLLM-MOD~\cite{videollm_mod} addresses this limitation by incorporating mixture of depth~\cite{mod} for efficient visual token computation, enabling higher visual input resolution. Flash-Vstream~\cite{flashvstream} and VideoStreaming~\cite{videostream} achieve real-time comprehension through a learnable memory module for stream compression. However, prior work has often lacked well-reasoned architectural designs and training strategies, consequently struggling to achieve a balance between efficiency and performance. Our approach introduces novel designs in both architectural structure and training strategy, leading to the development of a more powerful online video MLLM.

\noindent\textbf{Online Video Benchmarks.} 
VideoLLM-online~\cite{videollm_online} evaluates the model as an online video assistant on the streaming narration task.
MovieChat-1K~\cite{moviechat} introduces a breakpoint mode, which requires the model to ask and answer questions at different time points during video playback.
VStream-QA~\cite{flashvstream} represents the first benchmark to evaluate streaming multimodal video understanding. While it incorporates five types of timestamp-anchored questions generated through GPT-4 with human verification, its task paradigm largely mirrors offline scenarios. In contrast, \bench{} emphasizes real-time spatiotemporal detail comprehension in streaming contexts, featuring a comprehensive task set tailored to streaming video characteristics. It builds upon high-quality spatiotemporal understanding datasets through targeted refinement to ensure benchmark integrity and reliability.
\section{OVBench}\label{sec:ovbench}

\begin{table*}[]
  \centering
  \setlength\tabcolsep{8pt}
  \renewcommand{\arraystretch}{1.2} 
  \resizebox{1.0\textwidth}{!}{%
  \begin{tabular}{c|c|l}
  \hline
  \rowcolor[HTML]{E0E0E0} 
  \textbf{Task} & \textbf{Subtasks} & \textbf{Query Examples} \\ \hline
  
  \multirow{4}{*}{\begin{tabular}[c]{@{}c@{}}Temporal Hallucination \\ Verification \\(THV)\end{tabular}} 
  & Action Persistence & Is the person in the \text{[0.168, 0.193, 0.846, 0.996]} location in the current frame performing walking? \\ \cline{2-3}
  & Step Verification & Is the person still installing the motherboard right now? \\ \cline{2-3}
  & Object Presence & \begin{tabular}[c]{@{}l@{}}How many markers are there on the screen 14.0 seconds before? Does the number increase or decrease compared \\ with the past screen?\end{tabular} \\ \hline
  
  \multirow{3}{*}{\begin{tabular}[c]{@{}c@{}}\\Past Memory\\(PM)\end{tabular}} 
  & Action Retrieval & Where was the person currently performing the talk to (e.g., self, a person, a group) in the scene 8 seconds ago? \\ \cline{2-3}
  & Procedure Recall & Which step did the person perform for the longest duration in the last 60 seconds? \\ \cline{2-3}
  & Trajectory Retrieval & \begin{tabular}[c]{@{}l@{}}When does the sheep \text{[0.491, 0.386, 0.584, 0.615]} in the current screen first appear? Give the corresponding \\ position when it first appears.\end{tabular} \\ \hline

  \multirow{3}{*}{\begin{tabular}[c]{@{}c@{}}Future Prediction\\(FP)\end{tabular}} 
  & Action Anticipation & What action is the person currently in the \text{[0.328, 0.211, 0.436, 0.809]} location likely to do next? \\ \cline{2-3}
  & Goal/Step Prediction & My goal is 'make flower crown'. What are the next steps I should take? \\ \cline{2-3}
  & Movement Prediction & What direction do you think the baby \text{[0.0, 0.062, 0.526, 0.903]} may move towards in the next second? \\ \hline

  \multirow{2}{*}{\begin{tabular}[c]{@{}c@{}}Spatial Perception\\(SP)\end{tabular}} 
  & Action Location & What is the person at the location \text{[0.024, 0.122, 0.624, 0.979]} currently doing? \\ \cline{2-3}
  & Object Position & Which option most accurately describes the location of the blankets now? \\ \hline

  \multirow{4}{*}{\begin{tabular}[c]{@{}c@{}}Temporal Perception\\(TP)\end{tabular}} 
  & Action Sequence & What is the sequence of actions the person in the scene has performed recently? \\ \cline{2-3}
  & Step Localization & How long has the person in the scene been performing the 'restore the fixed battery components and the back cover'? \\ \cline{2-3}
  & Object Existence State & \begin{tabular}[c]{@{}l@{}}What is the time period the turtle \text{[0.459, 0.518, 0.501, 0.556]} in the current screen appears in the video? And what is \\ the time period in which it disappeared?\end{tabular} \\ \hline

  \multirow{3}{*}{\begin{tabular}[c]{@{}c@{}}Spatio-Temporal \\Perception\\(STP)\end{tabular}} 
  & Action Trajectory & \begin{tabular}[c]{@{}l@{}}What is the sequence of actions and the corresponding movement trajectory of the person currently in the \\ \text{[0.383, 0.304, 0.642, 0.991]} location?\end{tabular} \\ \cline{2-3}
  & Object Trajectory & \begin{tabular}[c]{@{}l@{}}What is the trajectory of the object among car \text{[0.482, 0.518, 0.485, 0.531]}, car \text{[0.561, 0.51, 0.616, 0.577]} in \\ the past 5 seconds, which moves the shortest distance?\end{tabular} \\ \hline
  \end{tabular}
  }
  \vspace{-0.3cm}
  \caption{Task examples of \bench{}. For simplicity, we selected only one question in each task's templates for the presentation. Complete template examples can be found in the Appendix.}
  \vspace{-0.3cm}
  \label{tab:task_examples}
  \end{table*}

In this section, we detail the development process of OV-Bench. Based on a foundational definition of temporal contexts, we first derive the task types in Figure~\ref{tab:task_examples} for online video streaming scenarios in section~\ref{sec:task}. We then introduce the detailed process of QA generation in Figure~\ref{fig:pipeline}. Examples are listed in  Table \ref{tab:task_examples}.

\subsection{Task Formulation}
\label{sec:task}

\textbf{Basic Temporal Context Definition.} 
 To systematically evaluate streaming video comprehension, we define three fundamental temporal contexts that characterize the relationship between a question's timestamp and the video timeline:
\textbf{(1) Current:} The temporal window focuses specifically on the exact frame at which the question is posed, potentially including a small number of preceding frames necessary for understanding the current state.
\textbf{(2) Past:} The sequence of frames preceding the question timestamp, containing historical information about actions, events, and object trajectories.
\textbf{(3) Future:} The sequence of frames following the question timestamp, capturing the subsequent events based on current actions and trajectories.

\noindent\textbf{Task Formulation.} Based on three core temporal contexts --- \textbf{Past ($\mathbf{P}$)}, \textbf{Current ($\mathbf{C}$)}, and \textbf{Future ($\mathbf{F}$)} --- we identify 6 essential capabilities for online models in streaming video scenarios, where "$\rightarrow $" represent the inference or verification process from one time period to another, "$\cup$" indicates a joint understanding of multiple periods:
 
\begin{itemize}
\item \textbf{Spatial Perception ($\mathbf{C}$):} Identify and quantify discrete actions in the current frame, describe positions and spatial relationships of objects.
\item \textbf{Temporal Perception ($\mathbf{C} \rightarrow \mathbf{P} \cup \mathbf{C}$):} Track sequences of actions extending into the present moment, assess the duration of ongoing events and determine the existence status of objects over previous frames.
\item \textbf{Spatio-temporal Perception ($\mathbf{C}\rightarrow\mathbf{P} \cup  \mathbf{C}$):} Provide a comprehensive description of object motion trajectories, detailing displacements and relative positions for single or multiple targets.
\item \textbf{Past Memory ($\mathbf{C}\rightarrow\mathbf{P};$ or $\mathbf{P}$):} Recall past events relevant to a given action, retrieve duration or goals achieved, or locate an object’s past position and status when queried.
\item \textbf{Temporal Hallucination Verification ($\mathbf{P} \leftrightarrow \mathbf{C}$):} Determine if an action observed in the past is still ongoing in the current frame, verify the state of events that have occurred, and analyze object location changes between past and current contexts.
\item \textbf{Future Prediction ($\mathbf{P} \cup  \mathbf{C} \rightarrow \mathbf{F}$):} Project likely upcoming actions based on observed motion patterns and current spatial-temporal configurations.
\end{itemize}

\subsection{QA Generation}
\label{sec:qa_pipeline}
\textbf{Data Collection.} 
Unlike previous online question-answering benchmarks~\cite{flashvstream}, which typically utilize LLMs to generate questions and answers, our task requires temporal and spatial detail understanding, where questions rely on specific timestamps and bounding box annotations for accurate spatiotemporal comprehension.
To comprehensively capture the dynamics of streaming video, we curated 8 datasets across 6 varied domains in Figure \ref{fig:pipeline}, each dataset is selected to align closely with the real-time demands of streaming video comprehension. We only select from their validation and test sets to prevent potential data leakage.

\begin{figure*}[ht!]
    \centering
    \includegraphics[width=0.95\textwidth, height=0.35\textwidth]{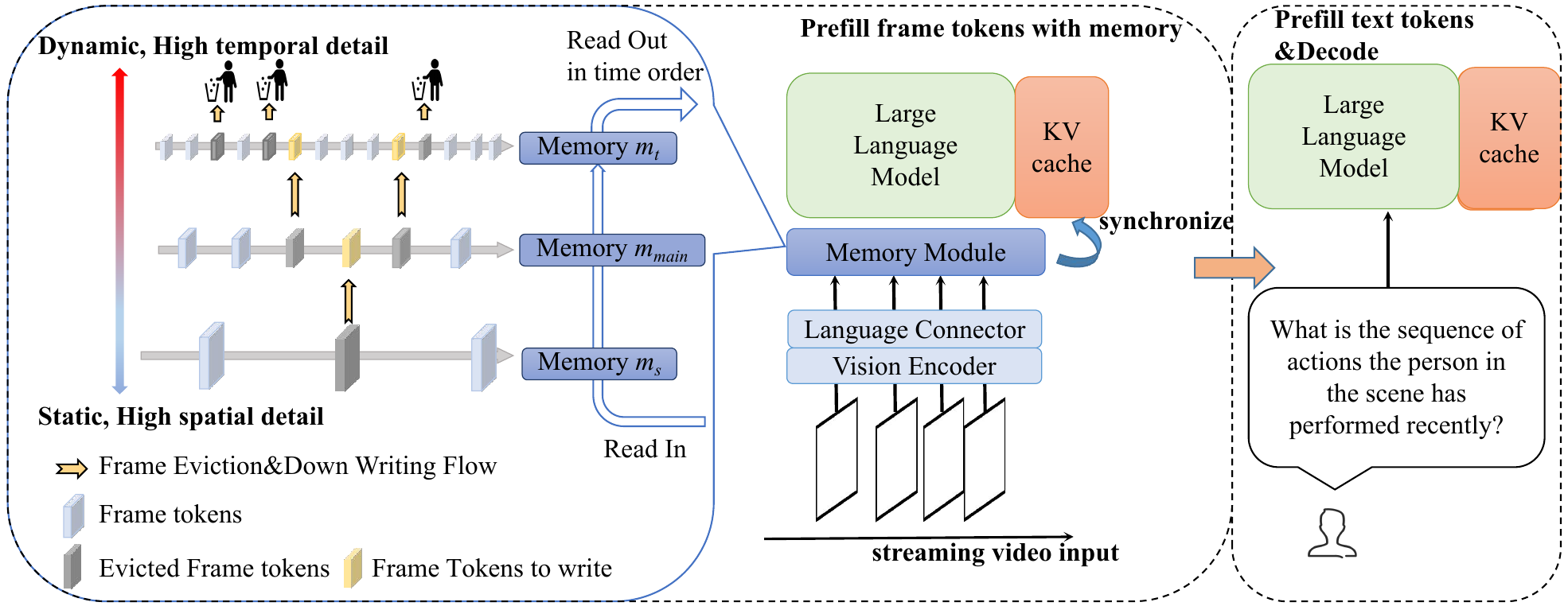}
    \caption{\textbf{Pyramid Memory Bank Architecture}: Illustrating the model's inference process with the pyramid memory bank structure. $m_{main}$ queues maintain balanced spatiotemporal information at different hierarchical levels,  $m_t$ is a high-frequency sampling queue for enhanced temporal detail preservation, and $m_s$ queue is for spatial detail retention. The system supports simultaneous frame input to both the memory bank and KVCache, with synchronization mechanisms for maintaining consistency during memory modifications.}
    \label{fig:memory}
\end{figure*}

\textbf{Option Generation for Streaming Video Scenes. }To ensure that answer options reflect the dynamic and shifting contexts in streaming video, we develop a multiple-choice generation process incorporating distractors that simulate real-world conditions. Distractors are selected from different timestamps within the same video, based on similar questions and objects or typical responses to such questions.

\textbf{Manual Check \& Sampling.}
\textbf{(1) Manual Check:} Our quality control involved several manual checks: question clarity and options ambiguity were assessed to avoid misinterpretations and annotation accuracy was verified by human annotators. 
\textbf{(2) Video Context Length Limitation:} Excessive video context length is trimmed according to the timestamp of the earliest relevant question. (i.e. The maximum time range that a question in the \textbf{Past Memory} task may be traced back to.)
\textbf{(3) Sampling:} Further, to ensure question diversity and balance, question distribution is optimized by scaling question count proportionally to video duration, ensuring diverse scene coverage, and maintaining task type balance. The appendix provides a detailed QA generation methodology across tasks, ensuring the transparency of our approach.

\section{Efficient Online Video Streams Modeling}

\subsection{Pyramid Memory Bank}

For online scenarios, as the number of input frames increases, it becomes essential to compress the visual tokens of the video in order to maintain real-time performance while preserving key information. Achieving a balance between spatial and temporal details within limited visual tokens is critical for effective spatiotemporal understanding.

To address this, we propose a \textbf{pyramid memory bank} structure that incrementally balances spatial and temporal details through progressive abstraction across multiple layers. As shown in Figure \ref{fig:memory}, the memory bank is divided into $n$ layers, denoted as $\{m_i \mid i=1, 2, \ldots, n\}$. Each layer progressively reduces spatial details in favor of temporal patterns by adjusting two key properties:

\textbf{Sampling Rate ($r_i$)}: Each layer $i$ samples frames from the input stream at a rate $r_i$, increasing progressively across layers to prioritize temporal continuity in deeper layers.

\textbf{Resolution} ($\text{Res}_i$): Each layer stores frames at a progressively lower resolution $\text{Res}_i$, ensuring that initial layers capture detailed spatial information, while deeper layers focus on temporal abstraction. The resolution for each layer is scaled as: $\text{Res}_i = \frac{\text{Res}_1}{\beta^{i-1}}$
  where $\text{Res}_1$ is the input frame resolution in the first layer, and $\beta > 1$ is a down-scaling factor. In practice, we use $\beta=2$.

Each memory layer $m_i$ performs 3 primary operations:

1. \textbf{Streaming Frame Writing}: The memory layer $m_i$ receives frames directly from the video stream, sampled according to $r_i$. These frames are stored in $m_i$ up to its capacity $C_i$. When the capacity $C_i $is full, perform the next operation.

2. \textbf{Frame Eviction\&Down Writing}: the memory layer identifies the most similar adjacent frame pair $\left(f_a^i, f_b^i\right)$, where cosine similarity is calculated after applying average pooling to each frame separately. The older frame in the pair is evicted, and its spatial information is reduced to the corresponding spatial scale $\text{Res}_{i+1}$ through average pooling before being passed to the next layer $m_{i+1}$:
   \begin{equation}
       f_{\text{next}}^{i+1} = \text{AvgPool2d}(f_{\text{evicted}}^{i}, \text{Res}_{i+1})
   \end{equation}

3. \textbf{Readout}: All stored frames across layers are read in temporal order when accessing memory banks.

\noindent\textbf{Compatibility with KVCache.} 
Existing memory-based compression methods, such as MovieChat~\cite{moviechat, moviechat+} and FlashVStream~\cite{flashvstream}, as the memory updates with each additional input frame, the entire compressed memory must be processed as a single unit when a user inputs frames, suffer from a bottleneck in compression efficiency. This all-at-once processing leads to memory compression as computational overhead, limiting real-time performance. In contrast, our memory bank aligns closely with KVCache, allowing frame tokens to be precomputed and stored efficiently. During \textbf{Frame Eviction} (operation 2), tokens after the timestamps of frames $f_a$ and $f_b$ are erased to maintain synchronization, as follows:
\begin{equation}
\text { KVCache } \leftarrow \text { KVCache } \backslash\left\{t_i \mid t_i>\min \left(t_{f_a}, t_{f_b}\right)\right\}
\end{equation}

where $t_{f_a}$ and $t_{f_b}$ denote the timestamps of frames $f_a$ and $f_b$, respectively. By erasing tokens after these timestamps, we can optimize both memory usage and real-time processing efficiency.

\subsection{Offline-to-Online Learning}

\begin{figure}[h!]
    \centering
    \includegraphics[width=0.95\linewidth, height=1.25\linewidth]{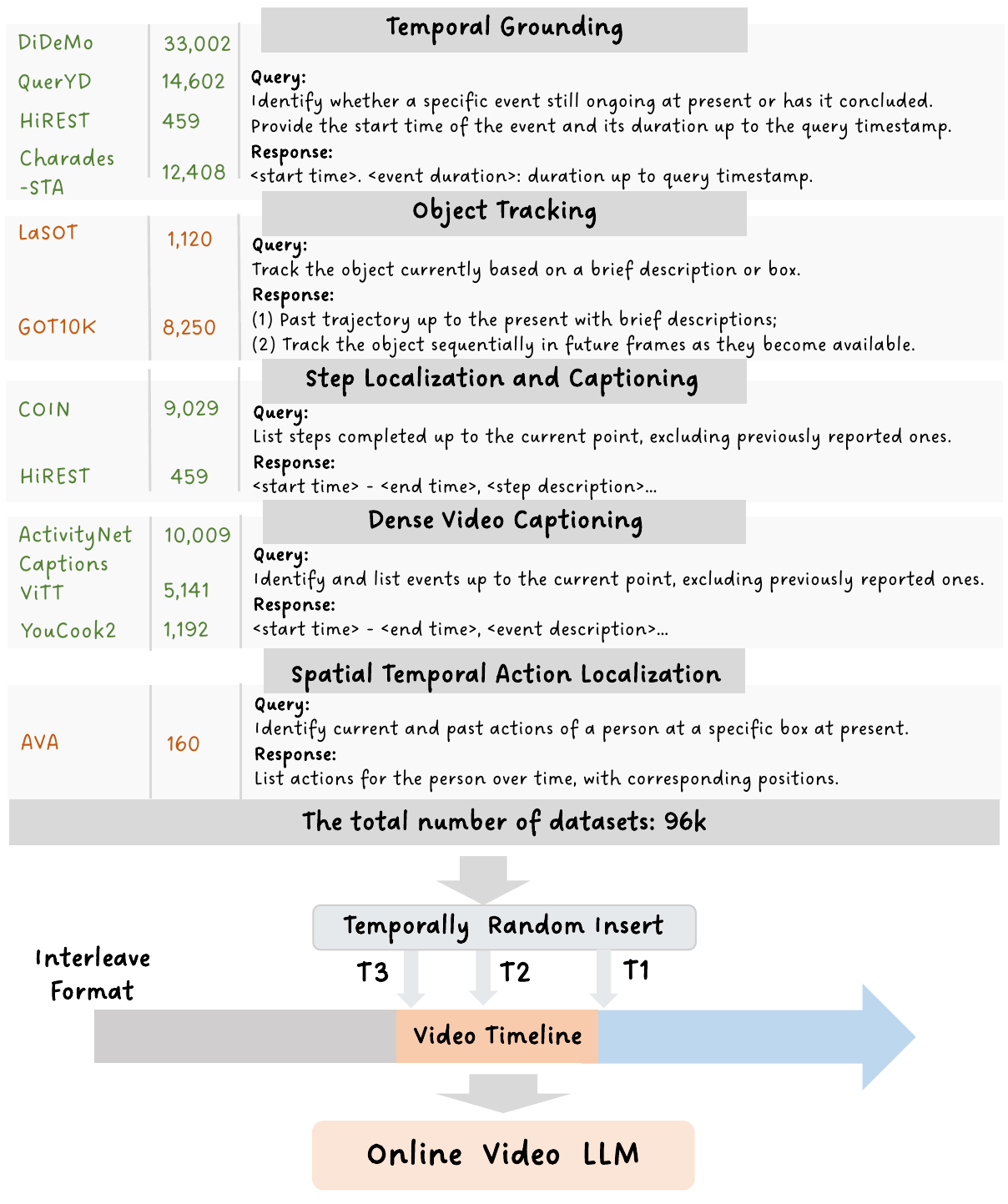}
    \caption{\textbf{Data Format Conversion Process for Online Spatiotemporal Instruction-Finetuning.} Our pipeline begins with 96K high-quality samples curated from 5 tasks across 12 datasets. The conversion process enhances online spatiotemporal understanding through template transformation. For each video sample, we strategically insert queries along the timeline in an organized interleaved format to facilitate temporal context differentiation.}
    \label{fig:data_format_conversion}
\end{figure}

\noindent\textbf{Data Collection.} 
To enhance the model's online spatiotemporal understanding capabilities, we prioritized datasets with rich spatiotemporal annotations. These datasets, featuring dense temporal annotations and spatial tracking information, inherently support multi-turn dialogue scenarios in streaming contexts.

\begin{itemize}
    \item \textbf{Fine-grained Event Temporal Boundary Identification.} To capture temporal event evolution, we leverage TimeChat-IT \cite{timechat} data, incorporating dense video captioning \cite{anet_dvc, youcook, vitt}, step localization \cite{coin, HiREST}, and temporal grounding datasets \cite{didemo, Charades-STA, queryd} for precise temporal boundary annotations.
    
    \item \textbf{Detailed Spatiotemporal Understanding.} We integrate object tracking \cite{lasot, got10k} and spatiotemporal action localization \cite{ava} annotations to enhance sequential object and action tracking capabilities, complementing the temporal information framework.
\end{itemize}

\noindent\textbf{Data Conversion.} 
To enhance multi-turn dialogue coherence and contextual awareness, we implement a structured temporal sampling strategy for question formulation, as shown in Figure \ref{fig:data_format_conversion}. Questions are positioned at specific temporal intervals while maintaining natural dialogue progression. Each sample's queries maintain task-category consistency to facilitate cross-dialogue temporal reasoning. Following MVBench \cite{mvbench}, we generate 5 diverse instructions per annotation task to ensure comprehensive interaction scenario coverage.

\noindent\textbf{Progressive Training.}
Optimizing fine-grained spatiotemporal understanding while maintaining timestamp and bounding box prediction capabilities during online training presents significant challenges. Inspired by curriculum learning~\cite{bengio2009curriculum}, we initially train the model on offline data to establish robust video understanding, followed by joint optimization with online data integration. An empirical analysis of this approach is presented in the ablation study section.

\begin{table*}[]
    \centering
    \resizebox{\textwidth}{!}{%
    \begin{tabular}{lllllllllllllllllll}
    \toprule 
    Task   Name                 & \multicolumn{1}{c}{}                                     & \multicolumn{3}{c}{FP}                                                  & \multicolumn{3}{c}{THV}                                                    & \multicolumn{3}{c}{PM}                                                  & \multicolumn{2}{c}{SP}                        & \multicolumn{2}{c}{STP}                         & \multicolumn{3}{c}{TP}                                                  & \multicolumn{1}{c}{}                      \\
    \cmidrule(lr){3-5} \cmidrule(lr){6-8} \cmidrule(lr){9-11} \cmidrule(lr){12-13} \cmidrule(lr){14-15} \cmidrule(lr){16-18}
    Subset Name                 & \multicolumn{1}{l}{\multirow{-2}{*}{Size}}           & \FPava & \FPevent & \FPtao & \Halava & \Halevent & \Haltao & \PMava & \PMevent & \PMtao & \SPava & \SPtao & \STPava & \STPtao & \TPava & \TPevent & \TPtao & \multicolumn{1}{c}{\multirow{-2}{*}{AVG}} \\
    \midrule
    \multicolumn{3}{l}{\cellcolor{gray!30}{\textit{sliding window size=32s fps=2}}}           & \cellcolor{gray!30}{}               & \cellcolor{gray!30}{}             & \cellcolor{gray!30}{}              & \cellcolor{gray!30}{}                & \cellcolor{gray!30}{}              & \cellcolor{gray!30}{}             & \cellcolor{gray!30}{}               & \cellcolor{gray!30}{}             & \cellcolor{gray!30}{}             & \cellcolor{gray!30}{}             & \cellcolor{gray!30}{}              & \cellcolor{gray!30}{}              & \cellcolor{gray!30}{}             & \cellcolor{gray!30}{}               & \cellcolor{gray!30}{}             &     \cellcolor{gray!30}{}                                      \\
    \midrule
    Gemini-1.5-Flash~\cite{gemini}           & \multicolumn{1}{c}{-}                                    & 71.4                  & 53.6                    & 21.9                  & 56.5                   & 60.8                     & 40.6                   & 36.7                  & 47.9                    & 62.5                  & 32.3                  & 37.5                  & 87.0                   & 50.0                   & 83.3                  & 22.3                    & 46.9                  & 50.7                                      \\
    \midrule
    InternVL2~\cite{chen2023internvl}                   & 7B                                                       & 52.6                  & 60.2                    & 27.6                  & 57.5                   & \cellcolor{gray!30}{52.0}                     & 58.5                   & \cellcolor{gray!30}{38.8}                  & \textbf{67.1}                    & 58.3                  & 38.1                  & \textbf{31.3}                  & 87.4                   & 37.0                   & \cellcolor{gray!30}{75.4}                  & \cellcolor{gray!30}{31.4}                    & 5.9                   & 48.7                                      \\
    InternVL2~\cite{chen2023internvl}                   & 4B                                                       & 57.7                  & 57.0                    & 14.4                  & 59.2                  & 49.4                     & \cellcolor{gray!30}{60.0}                   & 30.3                  & 61.8                    & 46.3                  & 30.9                  & 20.1                  & 83.0                   & 32.3                   & 70.7                  & 29.4                    & 3.4                   & 44.1                                      \\
    LLaMA-VID~\cite{llamavid}                   & 7B                                                       & 43.6                  & 50.9                    & 19.6                  & \cellcolor{gray!30}{64.0}                   & 47.5                     & 46.8                   & 29.4                  & 48.9                    & 51.2                  & 31.9                  & 11.2                  & 75.7                   & 24.8                   & 59.1                  & 26.0                    & \cellcolor{gray!30}{40.0}                  & 41.9                                      \\
    LLaVA-Onevision~\cite{LLaVA-OneVision}             & 7B                                                       & \textbf{68.0}                  & 62.7                    & \textbf{35.9}                  & 58.4                   & 50.3                     & 46.5                   & 29.4                  & 60.7                    & 58.0                  & \textbf{43.1}                  & 14.2                  & 86.5                   & \cellcolor{gray!30}{49.7}                   & 70.7                  & 28.1                    & 30.2                  & 49.5                                      \\
    LongVA~\cite{longva}                        & 7B                                                       & 64.1                  & 56.5                    & \cellcolor{gray!30}{29.5}                  & 54.9                   & 51.9                     & 34.8                   & 35.3                  & 55.6                    & 57.7                  & 31.6                  & 3.4                   & 67.4                   & 44.7                   & \textbf{80.0}                  & 26.7                    & 4.0                   & 43.6                                      \\
    MiniCPM-V2.6~\cite{hu2024minicpm}                & 7B                                                       & 33.3                  & 35.9                    & 15.0                  & 59.2                   & 50.8                     & 55.1                   & 25.0                  & 37.4                    & 41.7                  & 26.6                  & 11.8                  & \textbf{98.3}                   & 36.3                   & 66.1                  & 26.4                    & 6.2                   & 39.1                                      \\
    Qwen2-VL~\cite{wang2024qwen2vl}                   & 7B                                                       & \cellcolor{gray!30}{60.3}                  & \textbf{66.1}                   & 22.1                  & 54.9                   & 51.5                     & 51.1                   & 37.8                  & \cellcolor{gray!30}{64.4}                    & \cellcolor{gray!30}{69.3}                  & 35.3                  & 28.5                  & \cellcolor{gray!30}{97.0}                   & 49.4                   & 65.1                  & 30.8                    & 11.7                  & \cellcolor{gray!30}{49.7}                                      \\
    {LITA}~\cite{huang2025lita}                          & 7B                                                       &19.2	&24.5	&19.9	&40.8	&48.9	&24.9	&3.1	&27.3	&6.4	&6.9	&14.6	&35.2	&23.9	&27.4	&0.5	&3.4	&20.4

                     \\
    {TimeChat}~\cite{timechat}                    & 7B                                                       &7.7	&15.3	&18.7	&20.6	&15.7	&11.7	&9.1	&14.7	&9.8	&7.5	&19.5	&13.9	&10.3	&9.3	&10.1	&10.8	&12.8
                     \\
    VTimeLLM~\cite{huang2024vtimellm}                    & 7B                                                       & 37.2                  & 23.4                    & 15.0                  & \textbf{64.8}                   & 43.8                     & 53.2                   & 25.9                  & 38.8                 & 32.5                  & 25.9                  & 20.4                  & 40.9                   & 6.8                    & 48.4                  & \textbf{43.5}                    & 8.6                   & 33.1                                      \\
       \model~(Ours)                           & 4B                                                       & 56.4                  & \cellcolor{gray!30}{63.0}                    & 15.6                  & 57.1                   & \textbf{57.9}                     & \textbf{61.9}                   & \textbf{39.1}                  & 54.2                    & \textbf{73.9}                  & \cellcolor{gray!30}{41.3}                  & \cellcolor{gray!30}{29.7}                  & 92.2                   & \textbf{53.1 }                  & 69.8                  & 27.3                    & \textbf{69.9}                  & \textbf{53.9 }                                     \\
    \midrule
    \multicolumn{3}{l}{\cellcolor{gray!30}{\textit{Streaming video input at 2 fps.}}}             & \cellcolor{gray!30}{}               & \cellcolor{gray!30}{}             & \cellcolor{gray!30}{}              & \cellcolor{gray!30}{}                & \cellcolor{gray!30}{}              & \cellcolor{gray!30}{}             & \cellcolor{gray!30}{}               & \cellcolor{gray!30}{}             & \cellcolor{gray!30}{}             & \cellcolor{gray!30}{}             & \cellcolor{gray!30}{}              & \cellcolor{gray!30}{}              & \cellcolor{gray!30}{}             & \cellcolor{gray!30}{}               & \cellcolor{gray!30}{}             & \cellcolor{gray!30}{}                                 \\
    \midrule
    {VideoLLM-Online}~\cite{videollm_online}               & 7B                                                       &0	&1.8	&20.9	&5.2	&5.9	&32.6	&0	&2.3	&26.7	&0.6	&26.6	&0.9	&19.9	&0.9	&1.7	&8.3	&9.6
                     \\
    MovieChat~\cite{moviechat}                   & 7B                                                       & 23.1                  & 27.5                    & 23.6                  & 58.4                   & 43.9                     & 40.3                   & 25.6                  & 31.1                    & 23.9                  & 26.9                  & \textbf{39.6}                  & 24.4                   & 28.9                   & 29.3                  & 25.5                    & 21.9                  & 30.9                                      \\
    Flash-Vstream~\cite{flashvstream}                 & 7B                                                       & 26.9                  & 37.6                    & \textbf{23.9}                  & 60.1                   & 41.9                     & 40.0                   & 23.4                  & 35.3                    & 26.1                  & 24.7                  & 28.8                  & 27.0                   & 21.4                   & 29.8                  & 25.6                    & 26.8                  & 31.2                                      \\
  \model~(Ours)                        & 4B                                                       & \textbf{64.1}                  & \textbf{59.7}                    & 16.6                  & \textbf{63.1}                   & \textbf{58.3}                     & \textbf{62.8}                   & \textbf{42.2}                  & \textbf{54.4}                    & \textbf{70.6}                  & \textbf{54.1}                  & 24.8                  & \textbf{88.7}                   & \textbf{48.5}                   & \textbf{73.0}                  & \textbf{25.9}                    & \textbf{71.7}                  & \textbf{54.9}                                      \\
    \bottomrule
    
    \end{tabular}%
    }
\caption{Evaluations results on \benchname. Our 4B-parameter model demonstrates substantial performance advantages in two key comparisons: a 23.7\% improvement over existing streaming-capable models, and a 4.2\% enhancement compared to advanced offline MLLMs while maintaining deployment flexibility. For VideoLLM-Online, we modify the official script for evaluation on the \benchname. However, it cannot follow instructions accurately and generates either nothing or redundant information, see the appendix for more details.}
\vspace{-0.3cm}
\label{tab:ov_bench}
    \end{table*}
\section{Experiments}

\subsection{Implementation Details}

\noindent\textbf{Training Data.} To enhance the model's comprehensive video understanding, we supplemented the online training data with offline video data from VideoChat2-IT~\cite{videochat2}, STAR~\cite{wu2021star_situated_reasoning} and PerceptionTest~\cite{perception_test}, image data from ShareGPT4V~\cite{sharegpt4v}, ShareGPT4o~\cite{chen2023internvl}, as well as multi-image data from LLaVA-OneVision~\cite{LLaVA-OneVision}.

\noindent\textbf{Model Architecture.} We use InternVL2-4B~\cite{chen2023internvl} as a powerful baseline model for development, integrating InternViT-300M as the visual encoder and Phi-3~\cite{phi3} as the language model. 
The input frames of all training processes are obtained by sampling at 1 fps, and the maximum input frame number is controlled at 64 frames through uniform sampling. For inference, the video sample rate for each memory in the hierarchical memory bank is \{1, 2, 8\} with token per frame \{256, 64, 16\}, respectively. Maintain a consistent token ratio for each memory queue.

\noindent\textbf{Evaluation Settings.} As most multimodal large models currently cannot receive streaming video input, we adopted two distinct methods to evaluate the models effectively:

\begin{itemize}
    \item \textbf{Sliding Window Setting:} We perform a sliding window evaluation on advanced MLLMs, capturing a 32-second time window before the question-asking time and extracting frames at 2 fps. This method allows for temporal context while evaluating responses.
    
    \item \textbf{Streaming Setting:} In this setting, we input all video frames from the beginning of the clip up to the question timestamp, sampled at 2 fps, to evaluate the model's real-time performance.
\end{itemize}

\subsection{Main Results on \bench}

\begin{table}[]
\resizebox{0.47\textwidth}{!}{
\begin{tabular}{lccccccc}
\hline
\multicolumn{1}{c}{\multirow{2}{*}{Model}} &
  \multirow{2}{*}{Size} &
  \multirow{2}{*}{\begin{tabular}[c]{@{}c@{}}EgoSchema\end{tabular}} &
  \multirow{2}{*}{MLVU} &
  \multicolumn{2}{c}{VideoMME} &
  \multirow{2}{*}{MVBench} &
  \multirow{2}{*}{\begin{tabular}[c]{@{}c@{}}LongVideo\\ Bench\end{tabular}} \\ \cline{5-6}
\multicolumn{1}{c}{}                            &    &      &      & Overall & Long &      &      \\ \hline
Video-LLaVA~\cite{videollava}                 & 7B & 38.4 & 47.3 & 39.9    & 36.2 & -    & 39.1 \\
Chat-UniVi~\cite{chatunivi}              & 7B & -    & -    & 40.6    & 35.8 & -    & -    \\
LLaMA-VID~\cite{llamavid}                 & 7B & 38.5 & 33.2 & -       & -    & 41.9 & -    \\
TimeChat~\cite{timechat}                 & 7B & 33.0 & 30.9 & 34.7    & 32.3 & 38.5 & -    \\
MovieChat~\cite{moviechat}              & 7B & 53.5 & 25.8 & 38.2    & 33.4 & 55.1 & -    \\
Video-LLaMA2~\cite{videollama}         & 7B & 51.7 & 48.5 & 47.9    &      & 54.6 & -    \\
LLaVA-Next-Video~\cite{zhang2024llavanextvideo} & 7B & 43.9 & -    & 46.6    &      &      & 43.5 \\
ShareGPT4Video~\cite{chen2024sharegpt4video}    & 8B & -    & 46.4 & 39.9    & 35.0 & 51.2 & 39.7 \\
VideoChat2~\cite{mvbench}                 & 7B & 54.4 & 47.9 & 39.5    & 33.2 & 60.4 & 39.3 \\
LongVA~\cite{longva}                   & 7B & -    & 56.3 & 52.6    & 46.2 & -    & -    \\
Video-CCAM~\cite{fei2024video}                  & 9B & -    & 58.5 & 50.3    & 39.6 & 64.6 & -    \\
Video-CCAM~\cite{fei2024video}                  & 4B & -    & 56.5 & 50.1    & 40.9 & 62.8 & -    \\ \hline
\model~(Ours)                                            & 4B & \textbf{54.7} & \textbf{60.8} & \textbf{54.4}    & \textbf{47.1} & \textbf{65.2} & \textbf{54.1} \\
\bottomrule
\end{tabular}
}
\caption{VideoChat-Online's results on other offline long and short video benchmarks show comparable or better overall performance.}
\label{tab:offline_video_bench}
\end{table}

The main results on \bench \ are shown in Table \ref{tab:ov_bench}.

\noindent\textbf{Streaming Video LLMs Comparison.} 
In the streaming video setting, answering questions is more difficult because the model will be disturbed by more irrelevant context (see the FIFO and ``w/o compression" of Table \ref{tab:mem_update}, when the entire video from the beginning to the end frame is input, the performance is reduced by about 4.62\%).
Nevertheless, current online models have a non-negligible gap in real-time performance with existing offline models, suggesting that accessing rich, task-specific online data is crucial to narrowing this gap.
Although Flash-Stream~\cite{flashvstream} employs an online model architecture, training with offline data may introduce biases that impact real-time adaptability. VideoLLM-Online~\cite{videollm_online}, trained on streaming narration data and free-form dialogue data from a first-person perspective, faces challenges in generalizing across diverse contexts and applications. Comparisons of computational costs and online scenarios cases with other models can be found in the appendix.

\noindent\textbf{Offline MLLMs in a Sliding Window Setting.} Our results show that offline MLLMs are more effectively generalized to online tasks, as they can be seen as a special case of broader online scenarios (e.g., using fixed question timestamps at the end of the video rather than dynamically across the video). 
\noindent\textbf{Effective Knowledge Transfer from offline to online:} Offline models have demonstrated superior performance than native online models in online scenarios, highlighting that knowledge from offline models can successfully transfer to online applications. Developing and transforming the streaming video model architecture based on existing advanced MLLMs emerges as a superior choice. Furthermore, by incorporating limited online scene data, our model achieves optimal results with a small 4B LLM, outperforming Qwen2-VL by 4.19\%.

\subsection{Offline Benchmark Results}

We conduct experiments on offline video understanding benchmarks to comprehensively evaluate our model. The test videos range from a few seconds to one hour in length. As shown in Table~\ref{tab:offline_video_bench}, our model outperforms previous state-of-the-art approaches, achieving the highest scores across multiple benchmarks, including 54.7\% on Egoschema~\cite{egoschema}, 60.8\% on MLVU~\cite{zhou2024mlvu}, 54.4\% on VideoMME~\cite{videomme} (Overall), 47.1\% on VideoMME (Long), 65.2\% on MVBench~\cite{mvbench}, and 54.1\% on LongVideoBench~\cite{wu2024longvideobench}. This consistent performance across both online and offline scenarios demonstrates \model's robust generalization capabilities.

\subsection{Ablations of VideoChat-Online}
\begin{table}[]
\centering
\resizebox{0.45\textwidth}{!}{%
\begin{tabular}{c|ccc|cccc}
\toprule
\multirow{2}{*}{Memory Bank} & \multicolumn{3}{c|}{Memory Capacity} & \multicolumn{4}{c}{\benchname}                                   \\ \cline{2-8} 
                             & $m_t$             & $m_{main}$           & $m_s$   & SP            & TP            & STP           & Overall       \\ \hline
w/o $m_t$                       & 0              & 5            & 2    & 38.9          & 51.1          & 68.5          & 54.2          \\
w/o $m_{main}$                       & 20             & 0            & 2    & 35.4          & 55.7          & 69.2          & 54.3          \\
w/o $m_s$                       & 12             & 10           & 0    & 36.9          & \textbf{56.9} & \textbf{69.3} & 53.7          \\ \hline
Ours                         & 12    & 2   & 2    & \textbf{39.4} & \textbf{56.9} & 68.6          & \textbf{54.9}
\\ \bottomrule
\end{tabular}%
}
\caption{\textbf{Memory Structure Ablations.} We remove each memory module for evaluation but keep the fixed number of visual tokens.}
\label{tab:mem_struct}
\end{table}
\begin{table}[]
\centering
\resizebox{0.4\textwidth}{!}{%
\begin{tabular}{c|c|c}
\toprule
Update Policy                 & \benchname       & VideoMME-long \\ \hline
Token Merge~\cite{moviechat}     & 51.5          & 43.9          \\
First In First Out            & 54.0          &  41.3            \\
Uniform  Sample & 52.1          & 45.0          \\
w/o Compression & 49.3          &   \textit{oom}            \\ \hline
Ours            & \textbf{54.9} & \textbf{47.1}\\
\bottomrule
\end{tabular}%
}
\caption{\textbf{Memory Update Policies Ablations.} FIFO prioritizes recent data for real-time queries, while Uniform Sample is applied to offline models. The ``w/o Compression" method directly inputs raw video data without further processing. Our method outperforms existing approaches in both online and offline benchmarks.}
\label{tab:mem_update}
\end{table}


\noindent\textbf{Memory Bank Design.} We conduct comprehensive ablation studies on the memory component across two dimensions: 
\noindent\textbf{(1) Structure.} Table \ref{tab:mem_struct} demonstrates the necessity of each memory layer. We systematically evaluated the impact of removing each layer while maintaining computational parity by adjusting token allocations. We maintained computational equivalence by ensuring the total token count remained constant at 832 tokens across all configurations.
The results reveal distinct patterns: spatial-biased memory configurations significantly enhance SP performance (comparing ``w/o $m_t$"  with ``w/o $m_{main}$" and ``w/o $m_{s}$", while temporally distributed memory structures improve TP and STP metrics. Our final architecture achieves optimal spatiotemporal balance through strategic memory allocation.
\noindent\textbf{(2) Memory Updating Policy.} Table \ref{tab:mem_update} presents a comparative analysis of various memory bank update strategies, including Token Merge, FIFO queuing, uniform temporal sampling, and uncompressed frame input. While FIFO inherently prioritizes recent temporal information, our proposed strategy demonstrates superior performance by facilitating cross-temporal information interaction between memory banks. This advantage stems from addressing a fundamental limitation of MLLMs: their tendency to encode frames independently, where token similarity reflects relationships between frames. Simple merging operations fail to preserve crucial inter-frame temporal dynamics.

\begin{table}[]
\centering
\resizebox{0.45\textwidth}{!}{%
\begin{tabular}{ccc|c}
\hline
\multicolumn{3}{c|}{Training Strategy}                        & \multirow{3}{*}{AVG} \\ \cline{1-3}
\multicolumn{1}{c|}{\multirow{2}{*}{Offline Training}} & \multicolumn{2}{c|}{Online Training} &  \\ \cline{2-3}
\multicolumn{1}{c|}{}  & Progressive             & Interleaved &                      \\ \hline
\multicolumn{1}{c|}{} & \multicolumn{1}{c|}{}  &             & 44.12   \\
\multicolumn{1}{c|}{\checkmark} & \multicolumn{1}{c|}{}  &             & 45.23 (+1.11)               \\
\multicolumn{1}{c|}{\checkmark} & \multicolumn{1}{c|}{\checkmark} &             & 52.42 (+8.30)                \\
\multicolumn{1}{c|}{\checkmark} & \multicolumn{1}{c|}{}  & \checkmark           & 51.84 (+7.72)                \\  \hline
\multicolumn{1}{c|}{\checkmark} & \multicolumn{1}{c|}{\checkmark} & \checkmark           & \textbf{53.89 (+9.77)}                \\ \hline
\end{tabular}%
}
\caption{Ablation analysis on training strategy and data organization impact on model performance.}
\vspace{-0.3cm}
\label{tab:ablation_training_stage}
\end{table}
\begin{figure}
    \centering
    \includegraphics[width=\linewidth]{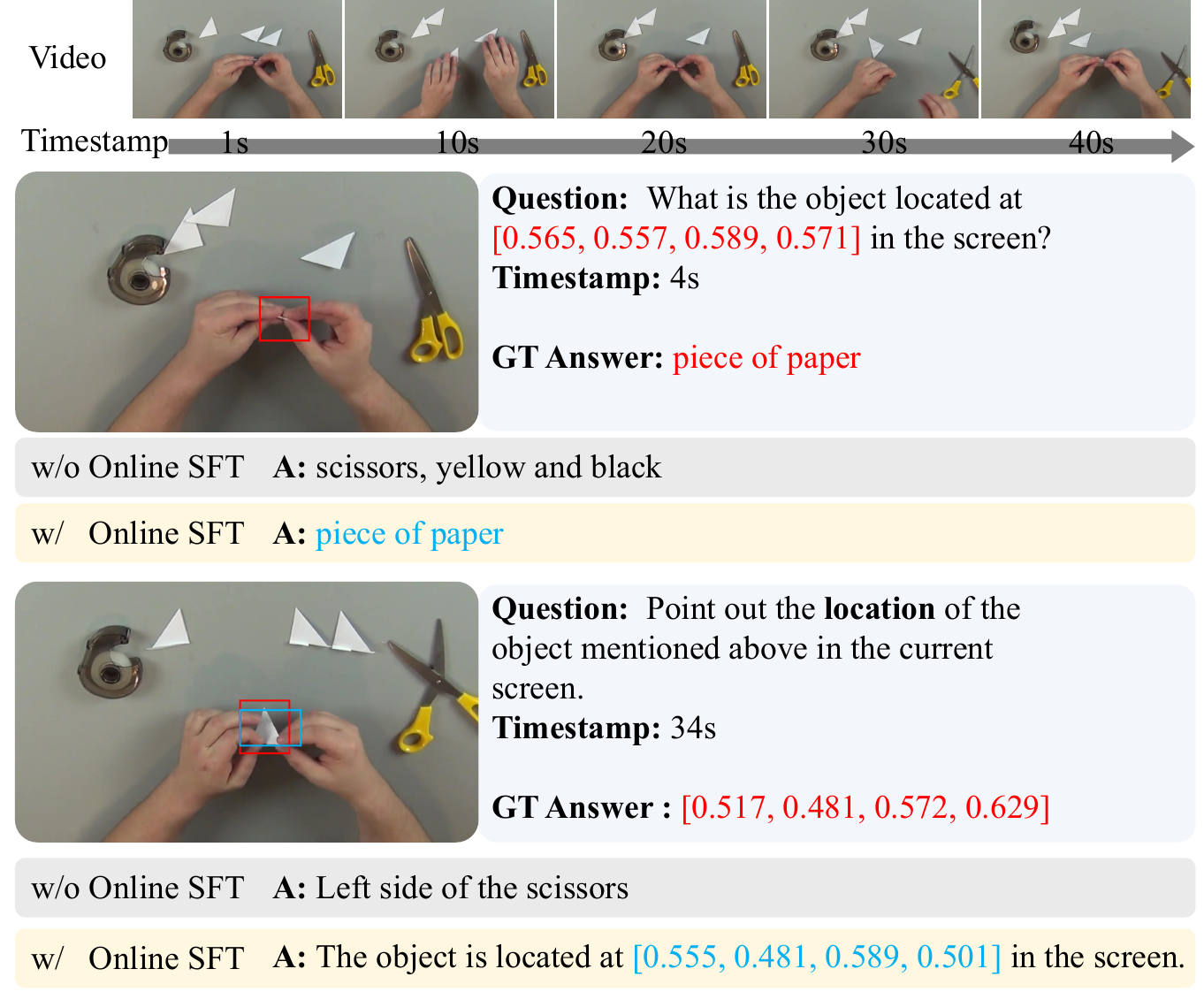}
    \caption{Qualitative comparison on online data training}
    \vspace{-0.3cm}
    \label{fig:aba_stream}
\end{figure}

\noindent\textbf{Impact of the Training Paradigm.} Table~\ref{tab:ablation_training_stage} presents a systematic analysis of different training strategies across three key dimensions:
\noindent\textbf{(1) Effect of Online Data:} Training exclusively with offline data yields a baseline performance of 45.23\%. Incorporating online data through progressive training elevates performance to 52.42\%, representing an 8.66\% improvement. Notably, this substantial enhancement is achieved with merely 96K online samples (6\% of the total training data), demonstrating the critical role of online data in developing temporal context understanding capabilities. 
\noindent\textbf{(2) Impact of Interleaved Data Format:} The integration of interleaved data format with offline training improves performance from 45.23\% to 51.84\%, with a 6.61\% increase. This enhancement suggests that interleaved data organization facilitates more effective learning of temporal relationships between question-answer pairs, particularly in dynamic online scenarios.
\noindent\textbf{(3) Progressive Training Strategy:} The combination of progressive training with interleaved format achieves 53.89\%, surpassing the joint training approach (51.84\%) by 2.05\%. This improvement indicates that transitioning from joint training to progressive online data introduction in the second epoch facilitates a better alignment progression from coarse to fine-grained spatiotemporal understanding.

\noindent\textbf{Qualitative comparison of Online SFT. } As shown in Figure~\ref{fig:aba_stream}, we verified it in a multi-round dialogue scenario and found that compared with the model based on offline training alone, the model after Online SFT accurately outputs the location and better builds the connection dialogue.

\section{Conclusion}
This work presents the following contributions to advance streaming video understanding: (1) \bench, a comprehensive benchmark designed to evaluate real-time spatiotemporal understanding capabilities; (2) \model, an efficient streaming video model that effectively balances efficiency. It achieves state-of-the-art performance while maintaining deployment flexibility. These advances provide a solid foundation for future research in streaming video understanding and real-world applications. 

\noindent\textbf{Acknowledgment}: This work is supported by the National Key R$\&$D Program of China (No. 2022ZD0160900), Jiangsu Frontier Technology Research and Development Program (No. BF2024076), the Collaborative Innovation Center of Novel Software Technology and Industrialization, and Nanjing University-China Mobile Communications Group Co., Ltd. Joint Institute.

{
    \small
    \bibliographystyle{ieeenat_fullname}
    \bibliography{main}
}

\clearpage
\setcounter{page}{1}
\maketitlesupplementary
\setcounter{section}{0}

\begin{figure*}[t!]
    \centering
    \includegraphics[width=1\linewidth]{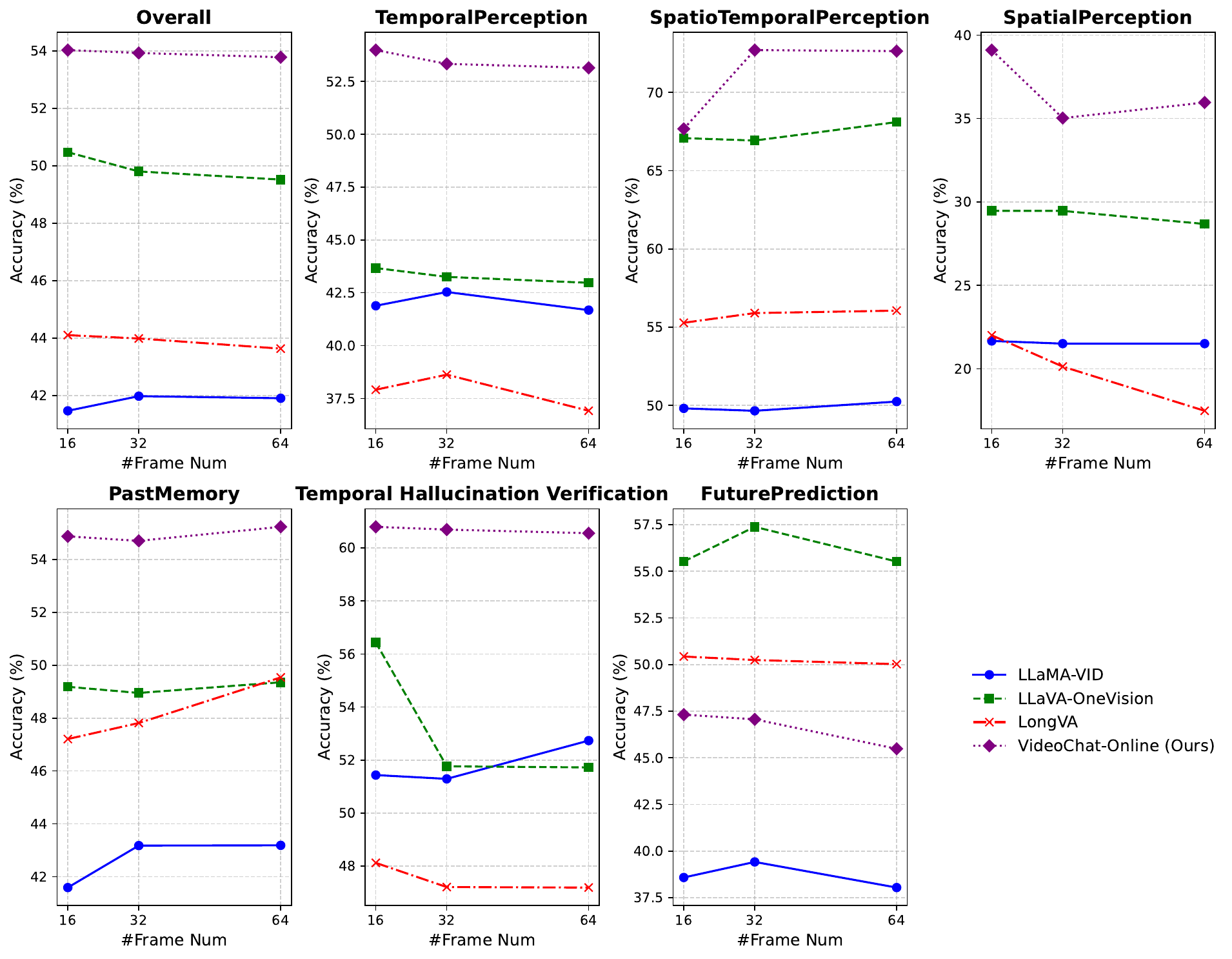}
    \caption{Models' performance in various subtasks and overall performance with varying input frames.}
    \label{fig:vary-frame}
\end{figure*}

\section{Implement Details for Ablation Study}

In this section, we describe the ablation experiments, focusing on the implementation details of comparison under different conditions.

\subsection{Memory Bank Design} 
The memory bank consists of three main modules: \textit{temporal memory} ($m_t$), \textit{main memory} ($m_{main}$), and \textit{spatial memory} ($m_s$). Each module stores a different number of frames and processes a distinct number of tokens per frame. The configuration is as follows:  
\begin{itemize}
    \item \textbf{Temporal memory ($m_t$)}: 12 frames, 16 tokens per frame.
    \item \textbf{Main memory ($m_{main}$)}: 2 frames, 64 tokens per frame.
    \item \textbf{Spatial memory ($m_s$)}: 2 frames, 256 tokens per frame.
\end{itemize}


\begin{table}[h!]
\centering
\begin{tabular}{l|c|c}
\toprule
\textbf{Model}       & \textbf{\benchname (\%)} & \textbf{VRAM Usage} \\ \midrule
InternVL2-4B~\cite{chen2023internvl}         & 44.1                  & \textit{oom}                 \\
MovieChat~\cite{moviechat}            & 30.9                  & 16.90 GB            \\
Flash-VStream~\cite{flashvstream}        & 31.2                  & 16.03 GB            \\ \midrule
\textbf{Ours}        & \textbf{54.9}         & \textbf{8.71 GB}    \\ \bottomrule
\end{tabular}%
\caption{Comparison of VARM evaluation results with state-of-the-art (SoTA) methods. Following the settings in Flash-VStream, we use 1000 video frames as input for VARM evaluation. Our method outperforms others in both \benchname accuracy and VRAM efficiency.}
\label{tab:vram}
\end{table}

\begin{table}[]
\centering
\begin{tabular}{l|c}
\toprule
Dataset                               & \benchname \\ \hline
Online Data SFT                       & 48.1    \\ \hline
w/o dense captioning                  & 47.0    \\
w/o step localization                 & 46.7    \\
w/o spatial temporal action detection & 44.8    \\
w/o temporal grounding                & 45.4    \\ \hline
\end{tabular}%
\caption{Ablation of the online instruction fine-tuning dataset by task category. For simplicity, we only use the online dataset for instruction fine-tuning for 1 epoch based on InternVL2-4B~\cite{chen2023internvl}.}
\label{tab:ablation-data}
\end{table}

\begin{table}[]
\centering
\begin{tabular}{ccc|c}
\hline
\multicolumn{3}{c|}{Memory Bank Capacity} & \benchname       \\ 
$m_t$      & $m_{main}$      & $m_s$      & Overall       \\ \hline
18         & 3               & 3          & 54.4          \\
24         & 4               & 4          & 54.4          \\ \hline
12         & 2               & 2          & \textbf{54.9} \\ \hline
\end{tabular}%
\caption{The impact of Memory Bank Capacity on performance}
\label{tab:supp_mem_cap}
\end{table}

\begin{table}[]
\centering
\resizebox{\linewidth}{!}{%
\begin{tabular}{l|l|c|c}
\hline
Vision Encoder                        & LLM BackBone        & Scale & \benchname       \\ \hline
\multirow{4}{*}{InternViT-300M-448px~\cite{chen2023internvl}} & Qwen2-0.5B-Instruct~\cite{qwen2} & 1B    & 44.6          \\
                                      & InternLM2-chat-1.8b~\cite{cai2024internlm2} & 2B    & 43.4          \\
                                      & Phi-3-mini-3.8B~\cite{phi3}     & 4B    & 44.1          \\
                                      & InternLM2.5-chat-7b~\cite{cai2024internlm2} & 8B    & \textbf{48.7} \\ \hline
\end{tabular}%
}
\caption{Performance of models at different scales}
\label{tab:ovbench_scale}
\end{table}
\begin{figure}
    \centering
    \includegraphics[width=\linewidth]{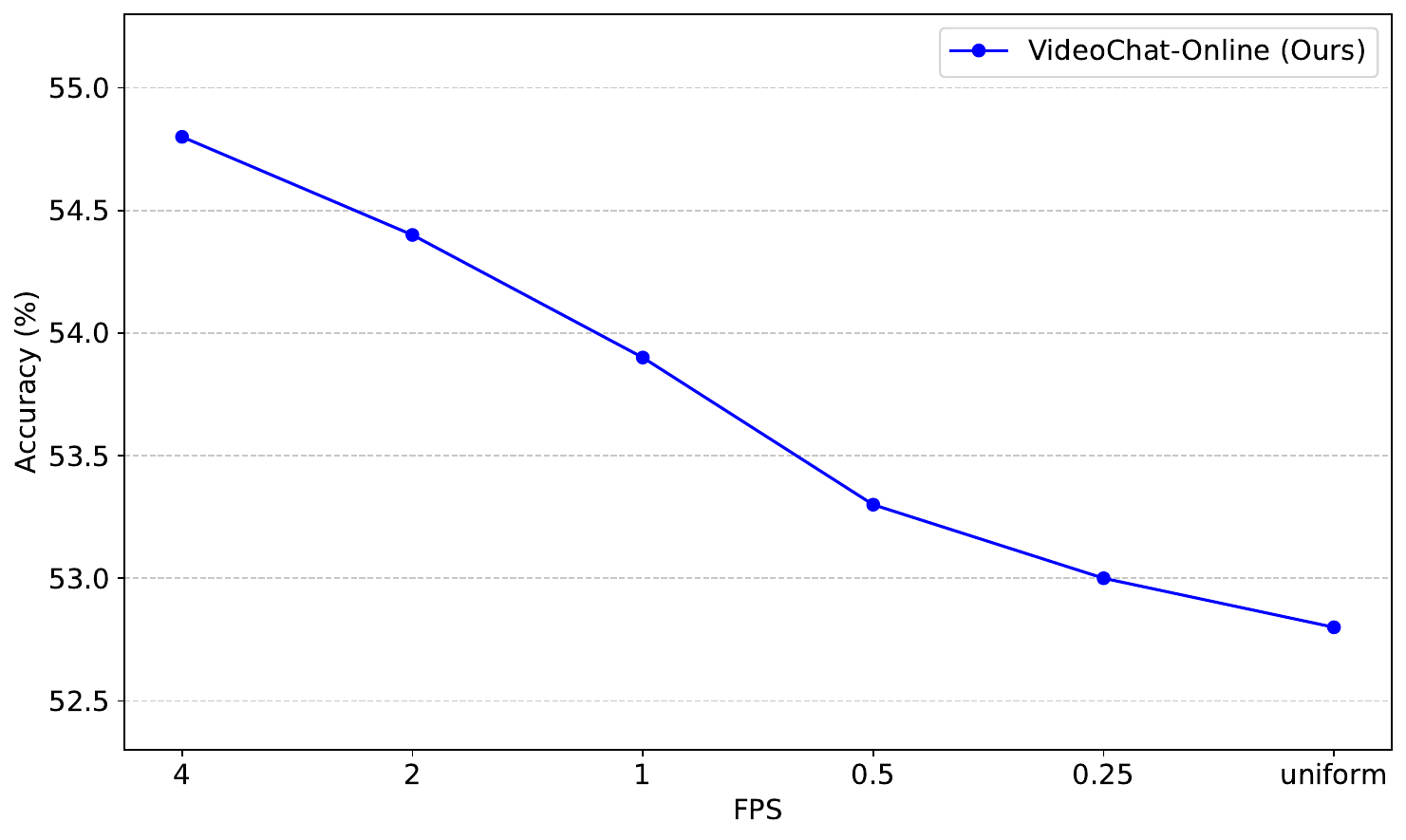}
    \caption{The impact of fps on model performance under the sliding window setting with 64 frames input.}
    \label{fig:vary-fps}
\end{figure}

\begin{figure*}
    \centering
    \includegraphics[width=\textwidth]{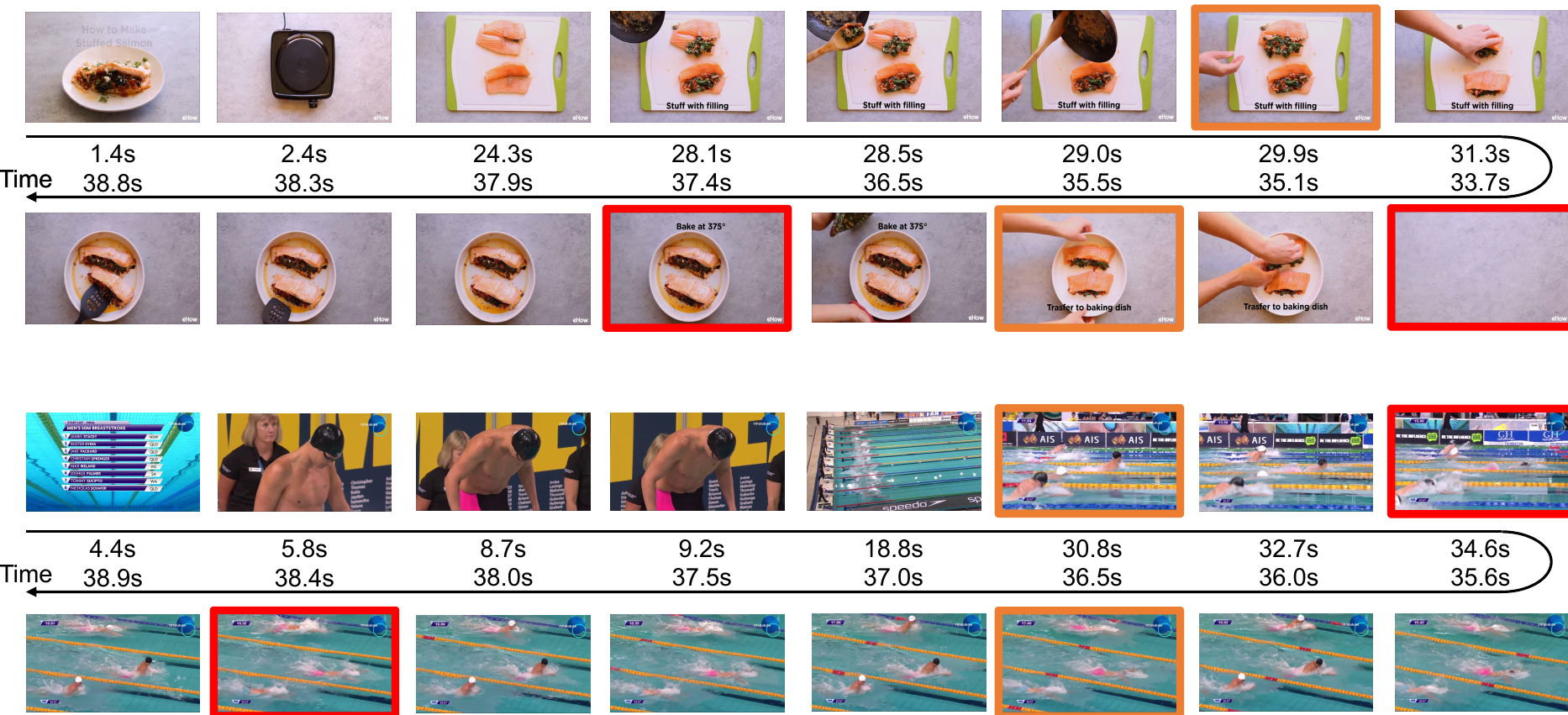}
    \caption{Visualization of the hierarchical memory bank. Frames in the $\textcolor{red}{m_s}$ layer are highlighted in \textcolor{red}{red}, those in the main layer $\textcolor{red}{m_{main}}$ are in \textcolor{orange}{orange}, and the remaining frames belong to the $\textcolor{blue}{m_t}$ layer. The structure illustrates the different capacities allocated to each layer.}
    \label{fig:memory_bank}
\end{figure*}

\begin{table*}[t]
\small
\begin{tabular}{p{3.5cm}|p{2.2cm}|p{2.5cm}|p{6.5cm}}
\toprule
\textbf{Task Categories} & \textbf{Source} & \textbf{Domain} & \textbf{QA Generation Protocol} \\
\midrule

• Action Discrepancy & AVA~\cite{ava} & Movie & \textbf{Question Requirements:} \\
• Action Localization & & & • Minimum 6 possible options available \\
• Action Retrieval & & & • Video context: max(900s, $t_{query}$ - 120s) \\
• Action Anticipation & & & • Continuous frame sequences only \\
• Action Sequence & & & \textbf{Answer Generation:} \\
• Action Trajectory & & & • Same video, different timestamps \\
& & & • Task-specific typical answers \\
& & & • Random select answers \\
\midrule
• Step Verification & HiREST~\cite{HiREST} & Instructional & \textbf{Question Requirements:} \\
• Procedure Recall & COIN~\cite{coin} & Indoor Activities & • Minimum 6 options available \\
• Goal/Step Prediction & & Open-Domain & • Video context: max(0s, $t_{query}$ - 300s) \\
• Step Localization & & & • Clear step descriptions only \\
& & & \textbf{Option Generation:} \\
& & & • Intra-video temporal alternatives \\
& & & • Similar topic cross-video options \\
& & & • Task-specific typical answers \\
& & & • Step Duration $\leq$ 5s \\
& & & • 3 $\leq$ Number of Steps $\leq$ 10 \\
\midrule

• Object Presence & TAO~\cite{TAO} & Road Scene & \textbf{Question Criteria:} \\
• Object Position & HACS~\cite{HACS} & Indoor Activities & • Specific object class labeling \\
• Trajectory Retrieval & ArgoVerse~\cite{ArgoVerse} & Outdoor Activities & • No ambiguous object class (e.g. \"maybe" \"unknown")\\
• Movement Prediction & BDD~\cite{BDD} & Open-Domain & \textbf{Answer Construction:} \\
• Object State & LaSOT~\cite{lasot} & & • Temporal consistency with question \\
• Object Trajectory & AVA~\cite{ava} & & • Class-consistent trajectories \\
& & & • if use template: 3×3 grid-based position mapping \\
& & & • Task-appropriate typical responses \\
\bottomrule
\end{tabular}
\caption{Task Categories and Question-Answer Generation Strategy}
\label{tab:qa_generation}

\end{table*}
\begin{table*}[h!]
\centering
\begin{tabular}{@{}p{0.25\textwidth}|p{0.7\textwidth}@{}}
\toprule
\textbf{Question at 9.0s} &
\textbf{What is the time period the pillow [0.725, 0.483, 0.991, 0.736] appears in the video? When does it disappear?} \\ 
\midrule
\textbf{Options:} &
(A) Appears: 2.0 - 4.0s, 6.0s, 8.0s; Disappears: 5.0s, 7.0s, 9.0s. \newline
(B) Appears: 1.0 - 4.0s, 6.0 - 8.0s; Disappears: 5.0s, 9.0s. \newline
(C) Appears: 1.0s; Disappears: 2.0 - 9.0s. \newline
(D) Appears: 7.0 - 9.0s. \\ 
\midrule
\textbf{Answer:} &
\textbf{Response:} Appears: 2.0 - 4.0s, 6.0s, 8.0s; Disappears: 5.0s, 7.0s, 9.0s. \newline
\textbf{Ground Truth:} D \newline
\textbf{Task Type:} TemporalPerception \\ 
\midrule
\textbf{Question at 17.0s} &
\textbf{When does the pillow [0.477, 0.443, 0.695, 0.61] first appear in the video? What is the position?} \\ 
\midrule
\textbf{Options:} &
(A) 8 seconds before: [0.391, 0.31, 0.587, 0.626]. \newline
(B) 2 seconds before: [0.375, 0.244, 0.472, 0.829]. \newline
(C) 10 seconds before: [0.855, 0.626, 1.0, 1.0]. \newline
(D) 25 seconds before: [0.354, 0.243, 0.691, 0.624]. \\ 
\midrule
\textbf{Answer:} &
\textbf{Response:} The pillow first appears at 8.391s. \newline
\textbf{Ground Truth:} C \newline
\textbf{Task Type:} PastMemory \\ 
\bottomrule
\end{tabular}%
\caption{More detailed information about the VideoLLM-Online test. It is not able to correctly generate answer options based on the questions, or the specific content related to the options.}
\label{tab:videollm-online}
\end{table*}

\begin{table*}[]
\centering
\resizebox{\textwidth}{!}{%
\begin{tabular}{lllll}
\hline
\multicolumn{5}{c}{\cellcolor[HTML]{ECF4FF}\textbf{Object Tracking}} \\ \hline
\multicolumn{1}{l|}{(t=60.0s) User} &
  \multicolumn{4}{l}{\begin{tabular}[c]{@{}l@{}}Using the description "tiger walking on the dry grass," track the object's past trajectory until\\ the current frame and then display each future frame sequentially.\end{tabular}} \\ \hline
\multicolumn{1}{l|}{(t=60.0s) Assistant} &
  \multicolumn{4}{l}{\begin{tabular}[c]{@{}l@{}}\textless{}Frames During 0.0s - 60.0s\textgreater\\ At 0.0s, {[}0.439, 0.349, 0.486, 0.443{]};\\ ...\\ At 33.0s, {[}0.619, 0.246, 0.672, 0.367{]};\\ At 34.0s, {[}0.634, 0.218, 0.69, 0.357{]};\\ ...\\ At 60.0s, {[}0.397, 0.206, 0.504, 0.576{]}.\end{tabular}} \\ \hline
\multicolumn{5}{l}{...} \\ \hline
\multicolumn{1}{l|}{(t=71.0s) User:} &
  \multicolumn{4}{l}{\textless{}Frame at 71.0s\textgreater{}} \\ \hline
\multicolumn{1}{l|}{(t=71.0s) Assistant:} &
  \multicolumn{4}{l}{At 71.0s, {[}0.403, 0.246, 0.497, 0.601{]}.} \\ \hline
\multicolumn{1}{l|}{(t=74.0s) User:} &
  \multicolumn{4}{l}{\textless{}Frames at 74.0s\textgreater{}} \\ \hline
\multicolumn{1}{l|}{(t=74.0s) Assistant:} &
  \multicolumn{4}{l}{At 74.0s, {[}0.279, 0.361, 0.52, 0.657{]}.} \\ \hline
\multicolumn{5}{c}{\cellcolor[HTML]{ECF4FF}\textbf{Dense Video Captioning}} \\ \hline
\multicolumn{1}{l|}{(t=81.5s) User:} &
  \multicolumn{4}{l}{\begin{tabular}[c]{@{}l@{}}\textless{}Frames During 0.0s - 80.0s\textgreater\\ Identify all new events that occurred and ended up to the current frame, \\ which have not been reported before. Provide their start times, durations, and descriptions\\  in the format: \textless{}start time\textgreater - \textless{}end time\textgreater (duration: \textless{}x\textgreater seconds), \textless{}description\textgreater{}.\end{tabular}} \\ \hline
\multicolumn{1}{l|}{(t=81.5s) Assistant:} &
  \multicolumn{4}{l}{\begin{tabular}[c]{@{}l@{}}28s - 54s (duration: 26.0 seconds), chop the lemon grass, ginger, onion, and chili finely.\\ 55s - 80s (duration: 25.0 seconds), preheat oil in a bowl and add onions, chili, lemon grass, \\ ginger, and garlic, stir, and put a lid on.\end{tabular}} \\ \hline
\multicolumn{1}{l|}{(t=161.9s) User:} &
  \multicolumn{4}{l}{\begin{tabular}[c]{@{}l@{}}\textless{}Frames During 81.5s - 161.9s\textgreater\\  (Similar query)\end{tabular}} \\ \hline
\multicolumn{1}{l|}{(t=161.9s) Assistant:} &
  \multicolumn{4}{l}{129s - 144s (duration: 15.0 seconds), add coconut milk, fish sauce, and soy sauce into the pan.} \\ \hline
\multicolumn{1}{l|}{(t=229s) User:} &
  \multicolumn{4}{l}{\begin{tabular}[c]{@{}l@{}}\textless{}Frames During 161.9s - 229.0s\textgreater\\ (Similar query)\end{tabular}} \\ \hline
\multicolumn{1}{l|}{(t=229s) Assistant:} &
  \multicolumn{4}{l}{220s - 228s (duration: 8.0 seconds), pour the sauce over the mussel shells.} \\ \hline
\multicolumn{5}{c}{\cellcolor[HTML]{ECF4FF}\textbf{Temporal Grounding}} \\ \hline
\multicolumn{1}{l|}{(t=16.9s) User:} &
  \multicolumn{4}{l}{\begin{tabular}[c]{@{}l@{}}Determine whether the event 'the person pours some water into the glass' is still happening \\ or has already finished. If ongoing, respond with: \textless{}start time\textgreater - now (duration: \textless{}x\textgreater seconds). \\ If completed, use: \textless{}start time\textgreater - \textless{}end time\textgreater (duration: \textless{}x\textgreater seconds).\end{tabular}} \\ \hline
\multicolumn{1}{l|}{(t=16.9s) Assistant:} &
  \multicolumn{4}{l}{The event happens from 7.3s to now (duration: 9.6 seconds).} \\ \hline
\multicolumn{5}{c}{\cellcolor[HTML]{ECF4FF}\textbf{Spatial Temporal Action Localization}} \\ \hline
\multicolumn{1}{l|}{(t=1061.0s) User:} &
  \multicolumn{4}{l}{\begin{tabular}[c]{@{}l@{}}\textless{}Frames During Clip Start - 1061.0s\textgreater\\ In the current frame, list each person's location and actions.\end{tabular}} \\ \hline
\multicolumn{1}{l|}{(t=1061.0s) Assistant:} &
  \multicolumn{4}{l}{\begin{tabular}[c]{@{}l@{}}In the current frame, person at {[}0.318, 0.334, 0.427, 0.504{]}: carry/hold (an object), walk.\\ person at {[}0.421, 0.235, 0.565, 0.444{]}: carry/hold (an object), run/jog.\end{tabular}} \\ \hline
\multicolumn{5}{c}{...} \\ \hline
\multicolumn{1}{l|}{(t=1134.0s) User:} &
  \multicolumn{4}{l}{What actions did the person at bbox {[}0.046, 0.002, 0.421, 0.996{]} perform now and earlier?} \\ \hline
\multicolumn{1}{l|}{(t=1134.0s) Assistant:} &
  \multicolumn{4}{l}{\begin{tabular}[c]{@{}l@{}}In the current frame, the person at {[}0.046, 0.002, 0.421, 0.996{]}: stand, listen to (a person), watch (a person).\\ At 1133.0s, the same person at {[}0.044, 0.004, 0.401, 0.996{]}: stand, carry/hold (an object), listen to (a person), watch (a person).\end{tabular}} \\ \hline
\multicolumn{5}{c}{\cellcolor[HTML]{ECF4FF}\textbf{Step Localization}} \\ \hline
\multicolumn{1}{l|}{(t=132.1s) User:} &
  \multicolumn{4}{l}{\begin{tabular}[c]{@{}l@{}}\textless{}Frames During 0.0s - 131.0s\textgreater\\ Based on the current observation, list the events that concluded up to the current frame \\ and were not reported in earlier outputs. Ensure your response follows the format: \\ \textless{}start time\textgreater - \textless{}end time\textgreater (duration: \textless{}x\textgreater seconds), \textless{}description\textgreater{}.\end{tabular}} \\ \hline
\multicolumn{1}{l|}{(t=132.1s) Assistant:} &
  \multicolumn{4}{l}{\begin{tabular}[c]{@{}l@{}}25.0s - 37.0s (duration: 12.0 seconds), install the new faucet.\\ 38.0s - 131.0s (duration: 93.0 seconds), connect the water pipe.\end{tabular}} \\ \hline
\multicolumn{1}{l|}{(t=182.0s) User:} &
  \multicolumn{4}{l}{\begin{tabular}[c]{@{}l@{}}\textless{}Frames During 132.0s - 179.0s\textgreater\\ (Similar query)\end{tabular}} \\ \hline
\multicolumn{1}{l|}{(t=182.0s) Assistant:} &
  \multicolumn{4}{l}{132.0s - 179.0s (duration: 47.0 seconds), open the sluice and test the new faucet.} \\ \hline
\end{tabular}%
}
\caption{Instruction template examples and formatted output answers for each task.}
\label{tab:my-table}
\end{table*}
\begin{table*}[]
\centering
\resizebox{1\textwidth}{!}{%
\begin{tabular}{ccl}
\hline
\rowcolor[HTML]{C0C0C0} 
Temporal Context &
  Spatial Context &
  Query Examples \\ \hline
\multicolumn{1}{c|}{} &
  \multicolumn{1}{c|}{Action Discrepancy} &
  1) Is the person in the {[}0.168, 0.193, 0.846, 0.996{]} location in the current frame performing the walk? \\ \cline{2-3} 
\multicolumn{1}{c|}{} &
  \multicolumn{1}{c|}{Step Verify} &
  1) Is the person in the current frame still performing the 'install the motherboard'? \\ \cline{2-3} 
\multicolumn{1}{c|}{\multirow{-3}{*}{\begin{tabular}[c]{@{}c@{}}Temporal Hallucination\\ Verification\end{tabular}}} &
  \multicolumn{1}{c|}{Object Presence} &
  \begin{tabular}[c]{@{}l@{}}1) Is the umbrella {[}0.507, 0.606, 0.612, 0.868{]} still in the screen 3.0 seconds before? \\ 2) How many markers are there on the screen 14.0 seconds before? Does the number increase or decrease \\ compared with the past screen?\end{tabular} \\ \hline
\multicolumn{1}{c|}{} &
  \multicolumn{1}{c|}{Action Location} &
  \begin{tabular}[c]{@{}l@{}}1) What action is the person at the location {[}0.024, 0.122, 0.624, 0.979{]} currently performing?\\ 2) How many people in the current frame are performing the action:  carry/hold (an object) ?\\ 3) Where is the person currently performing the talk to (e.g., self, a person, a group)  located in the picture?\end{tabular} \\ \cline{2-3} 
\multicolumn{1}{c|}{\multirow{-5}{*}{SpatialPerception}} &
  \multicolumn{1}{c|}{Object Position} &
  \begin{tabular}[c]{@{}l@{}}1) Based on visible information, which option most accurately describes the location of the blankets on the screen? \\ (Note: Positions with counts, e.g., 'left-middle (2) ', indicate multiple objects in the same area.) \\ 2) Which option most accurately describes the relative positions of other sheep with respect to \\ the reference position {[}0.388, 0.288, 0.509, 0.51{]} on the screen?\end{tabular} \\ \hline
\multicolumn{1}{c|}{} &
  \multicolumn{1}{c|}{Action Retrieval} &
  \begin{tabular}[c]{@{}l@{}}1) Where was the person currently performing the talk to (e.g., self, a person, a group)  in the scene 8 seconds ago? \\ 2) How many people were performing the watch (a person)  in the scene 60 seconds ago?\end{tabular} \\ \cline{2-3} 
\multicolumn{1}{c|}{} &
  \multicolumn{1}{c|}{Procedure Recall} &
  \begin{tabular}[c]{@{}l@{}}1) What goal was achieved in this video? \\ 2) Did the person follow the correct procedure to achieve the 'wash dish'? \\ 3) What actions did the person perform in sequence in the last 90 seconds?\\ 4) What steps did the person not perform in the last 15 seconds? \\ 5) How long has the person been performing the 'drive the car backward' in the last 90 seconds?\\ 6) Which action did the person perform for the longest duration in the last 15 seconds?\\ 7) What actions was the person performing before the last 30 seconds?\end{tabular} \\ \cline{2-3} 
\multicolumn{1}{c|}{\multirow{-10}{*}{PastMemory}} &
  \multicolumn{1}{c|}{Trajectory Retrieval} &
  \begin{tabular}[c]{@{}l@{}}1) Where is the location of the monkey {[}0.516, 0.49, 0.679, 0.804{]} on the screen 17.0 seconds before?\\ 2) When does the sheep {[}0.491, 0.386, 0.584, 0.615{]} in the current screen first appear in the video? \\ Give the corresponding position when it first appears.\end{tabular} \\ \hline
\multicolumn{1}{c|}{} &
  \multicolumn{1}{c|}{Action Anticipation} &
  \begin{tabular}[c]{@{}l@{}}1) What action is the person currently in the {[}0.328, 0.211, 0.436, 0.809{]} location likely to do next? \\ 2) What location in the frame is the person currently in the {[}0.485, 0.386, 0.578, 0.7{]} location likely to move to next?\end{tabular} \\ \cline{2-3} 
\multicolumn{1}{c|}{} &
  \multicolumn{1}{c|}{Goal/Step Prediction} &
  \begin{tabular}[c]{@{}l@{}}1) My goal is 'make flower crown'. What are the next steps I should take? \\ 2) Based on the series of actions performed by the person in the video, what is the ultimate goal?\end{tabular} \\ \cline{2-3} 
\multicolumn{1}{c|}{\multirow{-5}{*}{FuturePrediction}} &
  \multicolumn{1}{c|}{Movement Prediction} &
  1) What direction do you think the baby {[}0.0, 0.062, 0.526, 0.903{]} may move towards in the next second? \\ \hline
\multicolumn{1}{c|}{} &
  \multicolumn{1}{c|}{Action Sequence} &
  1) What is the sequence of actions the person in the scene has performed recently? \\ \cline{2-3} 
\multicolumn{1}{c|}{} &
  \multicolumn{1}{c|}{Step Localization} &
  1) How long has the person in the scene been performing the 'restore the fixed battery components and the back cover'? \\ \cline{2-3} 
\multicolumn{1}{c|}{\multirow{-3}{*}{TemporalPerception}} &
  \multicolumn{1}{c|}{Object Existence State} &
  \begin{tabular}[c]{@{}l@{}}1) What is the time period the turtle {[}0.459, 0.518, 0.501, 0.556{]} in the current screen appears in the video? \\ And what is the time period in which it disappeared?\end{tabular} \\ \hline
\multicolumn{1}{c|}{} &
  \multicolumn{1}{c|}{Action Trajectory} &
  1) What is the sequence of actions and the corresponding movement trajectory of the person currently in the {[}0.383, 0.304, 0.642, 0.991{]} location? \\ \cline{2-3} 
\multicolumn{1}{c|}{\multirow{-2}{*}{SpatioTemporalPerception}} &
  \multicolumn{1}{c|}{Object Trajectory} &
  \begin{tabular}[c]{@{}l@{}}1) What is the trajectory of the object among car {[}0.482, 0.518, 0.485, 0.531{]}, car {[}0.561, 0.51, 0.616, 0.577{]} in the past 5 seconds, \\ which moves the shortest distance? If an object disappears in the middle, calculate the distance based on the time period it last appears. \\ 2) In the video, what is the trajectory of the person {[}0.049, 0.103, 1.0, 1.0{]} in the past 2 seconds? Also, point out the period it disappears.\\ 3) Compared with 5 seconds ago, are the person {[}0.295, 0.614, 0.372, 1.0{]} and the guitar {[}0.299, 0.712, 0.419, 0.847{]} closer or farther apart?\\ 4) What is the trajectory of the object among person {[}0.315, 0.258, 0.671, 1.0{]}, nutcracker {[}0.322, 0.768, 0.487, 1.0{]} in the past 3 seconds, \\ which moves the shortest distance? If an object disappears in the middle, calculate the distance based on the time period it last appears.\end{tabular} \\ \hline
\end{tabular}%
}
\caption{Task Hierarchy and Question Templates: Overview of task categories, their subcategories, and corresponding example question templates. Each task is designed to probe specific spatiotemporal reasoning capabilities in video understanding, ranging from hallucination detection to future action prediction.}
\label{tab:task_templates}
\end{table*}
\begin{table}[]
\centering
\begin{tabular}{ll}
\hline
\textbf{Hyper-parameter}           & \textbf{Value}                                        \\ \hline
\multicolumn{2}{l}{Visual Encoder}                 \\ \hline
Frame Sampling Rate                & 1 FPS                                                 \\
Max Frames                         & 64                                                 \\
Preprocessing                      & Center Crop                                           \\
Input Resolution                   & 448 $\times$ 448                                      \\
Patch Size                         & 14 $\times$ 14 \\
Trainable?                         & False 

\\ \hline
\multicolumn{2}{l}{Frame Compressor}               \\ \hline
Pixel shuffle scale factor    & 0.5    \\
AvgPool2d Output Size    & \{16$\times$16, 8$\times$8, 4$\times$4\}    \\
 \hline
\multicolumn{2}{l}{MLP Projector} \\ \hline
Number of Layers                   & 2                                                     \\
Hidden Size                        & 4096                                                  \\
Output Size                        & 3072                                                  \\ 
Trainable?                         & True \\
\hline
\multicolumn{2}{l}{Large Language Model} \\ \hline
Architecture                        & Phi-3~\cite{phi3} \\
Trainable?                         & True \\
\hline

\multicolumn{2}{l}{Model Training}                       \\ \hline
Offline Training Epochs                   & 1                                                     \\
Online Joint Training Epochs                   & 1                                                     \\
Batch Size                         & 1024                                                    \\
Learning Rate             & 1e-4                                                  \\
Weight Decay                       & 0.05                                                   \\
Warmup Ratio                       & 0.03                                                  \\
LR Scheduler Type                  & Cosine                                                \\
Optimizer                          & AdamW \cite{adamw} \\
AdamW $\beta_1$, $\beta_2$         & (0.9, 0.999)  \\      \bottomrule                                 
\end{tabular}%
\caption{Hyper-parameters for fine-tuning.}
\label{tab:train_hype}
\end{table}
\begin{table}[]
\centering

\begin{tabular}{ll}
\hline
\textbf{Memory Bank}           & \textbf{Value}                                        \\ \hline
\multicolumn{2}{l}{Frame Sampling Rate}                 \\ \hline
$m_s$                & 1 FPS \\
$m_{main}$                & 2 FPS \\
$m_t$                & 8 FPS \\ 
\hline
\multicolumn{2}{l}{Capacity for Online Benchmark (Token Per Frame $\times$ Frames)} \\ \hline
$m_s$                & 256\ tokens $\times$ 2 Frames \\
$m_{main}$           & 64 tokens $\times$ 2 Frames\\
$m_t$                & 16 tokens $\times$ 12 Frames \\
Total Tokens         & 832 tokens \\
\hline
\multicolumn{2}{l}{Capacity for Offline Benchmark} \\ \hline
$m_s$                & 256\ tokens $\times$ 24 Frames \\
$m_{main}$           & 64 tokens $\times$ 24 Frames\\
$m_t$                & 16 tokens $\times$ 144 Frames \\
Total Tokens         & 9984 tokens \\
\hline
\end{tabular}%
\caption{Pyramid Memory Bank Hyper-parameters for Inference.}
\label{tab:infer_hype}

\end{table}

\begin{figure}
    \centering
    \includegraphics[width=\linewidth]{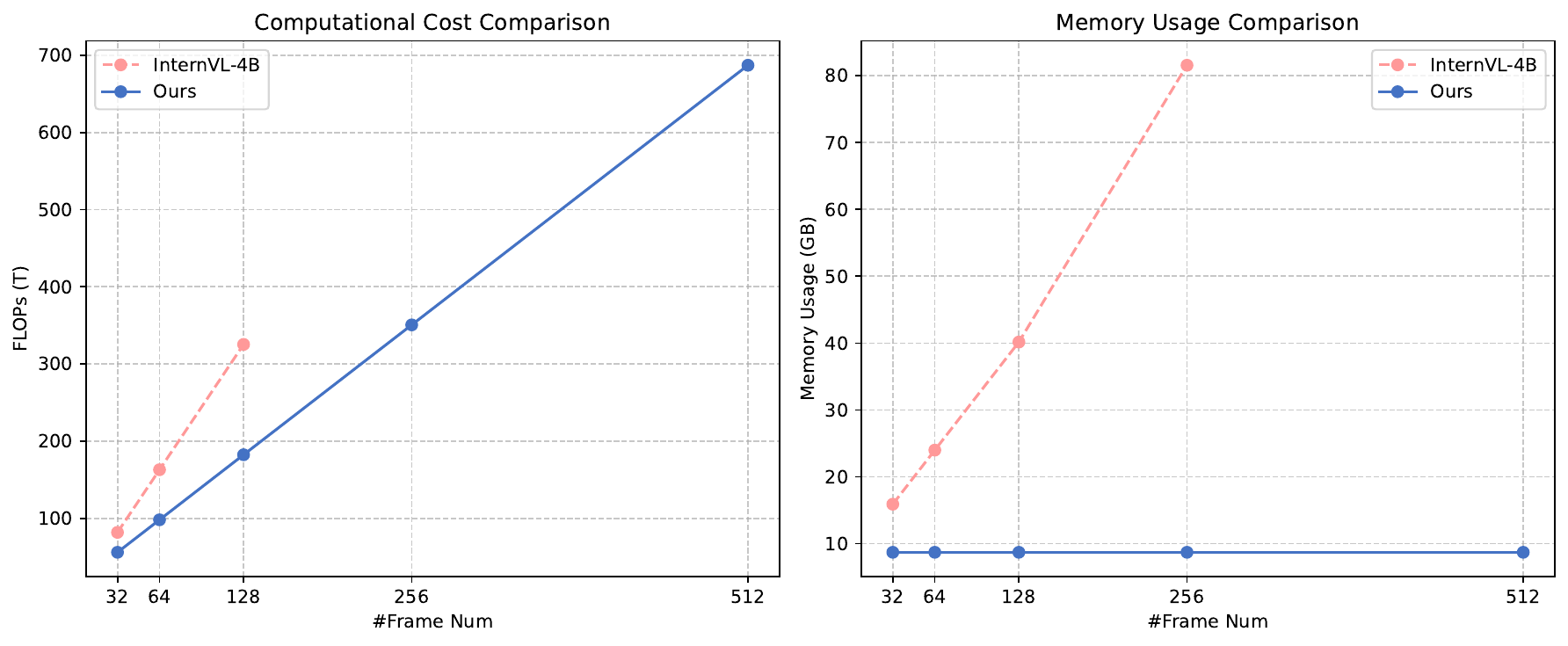}
    \caption{Comparison of computational cost and memory usage between baseline model (InternVL2-4B~\cite{chen2023internvl}) and our method.}
    \label{fig:computation_cost}
\end{figure}

\noindent\textbf{Total Computational Overhead:}  
The total computational overhead for processing all memory modules is \textbf{832 tokens}, calculated as:  
\[
\text{Total Tokens} = (12 \times 16) + (2 \times 64) + (2 \times 256) = 832 \text{ tokens}.
\]

This setup represents the \textbf{baseline model}. Subsequent experiments evaluate the impact of removing each memory module and redistributing the computational load to the remaining ones while maintaining the same overall computational budget.

\noindent\textbf{Conditions for Removing Memory Modules:}  
\begin{itemize}
    \item \textbf{w/o $m_t$}: Temporal memory is removed, and the load is shifted to the main memory.
    \item \textbf{w/o $m_{main}$}: Main memory is removed, and the load is redistributed to the temporal memory.
    \item \textbf{w/o $m_s$}: Spatial memory is removed, and the load is transferred to the main memory.
\end{itemize}

\subsection{Memory Updating Policy} 
In this section, we present the implementation details of different baselines in the Update Policy.

\textbf{Token merge:} from MovieChat~\cite{moviechat}: When the capacity of any memory module is full, the adjacent frames with the greatest similarity are merged.

\textbf{FIFO:} When any memory module is full, the frame with the earliest timestamp is evicted.

\textbf{Uniform Sample}: The video clips ending at the current problem timestamp are uniformly sampled, and the number of frames corresponds to the inherent capacity of each memory module.

\textbf{w/o Compression}: No memory compression, input all frames at fps=2.

\subsection{Training Paradigm} 
We evaluate the training strategy under the sliding window setting without introducing a memory bank for simplicity.

\noindent\textbf{Progressive training strategy.} If the progressive training strategy is adopted, the online data is introduced in the second epoch for joint training with online data, otherwise, the online data is introduced in the first epoch for joint training.
Compared with the performance obtained by directly using joint training (51.84\%), the performance obtained by progressive training (53.89\%) is significantly improved (+2.05\%).

\noindent\textbf{Non-interleaved data organization.} we train each query as an independent QA sample pair in the original interleaved form and keep the rest of the settings unchanged. The performance obtained by training with interleaved data (53.89\%) is better than that obtained by training with non-interleaved data (52.42\%), an improvement of 1.47\%.

\section{More Benchmark Results}

\textbf{For the VideoLLM-Online's evaluation,} we provide more detailed results in Table~\ref{tab:videollm-online}. It cannot correctly generate answer options based on the questions, or the specific content related to the options.

\textbf{Efficiency comparison.} We compared the efficiency with our baseline model InternVL2-4B in Figure~\ref{fig:computation_cost} and the existing state of art model in Table~\ref{tab:vram}, highlighting the efficiency advantages of our model.

\begin{figure*}
    \centering
    \includegraphics[width=1\linewidth]{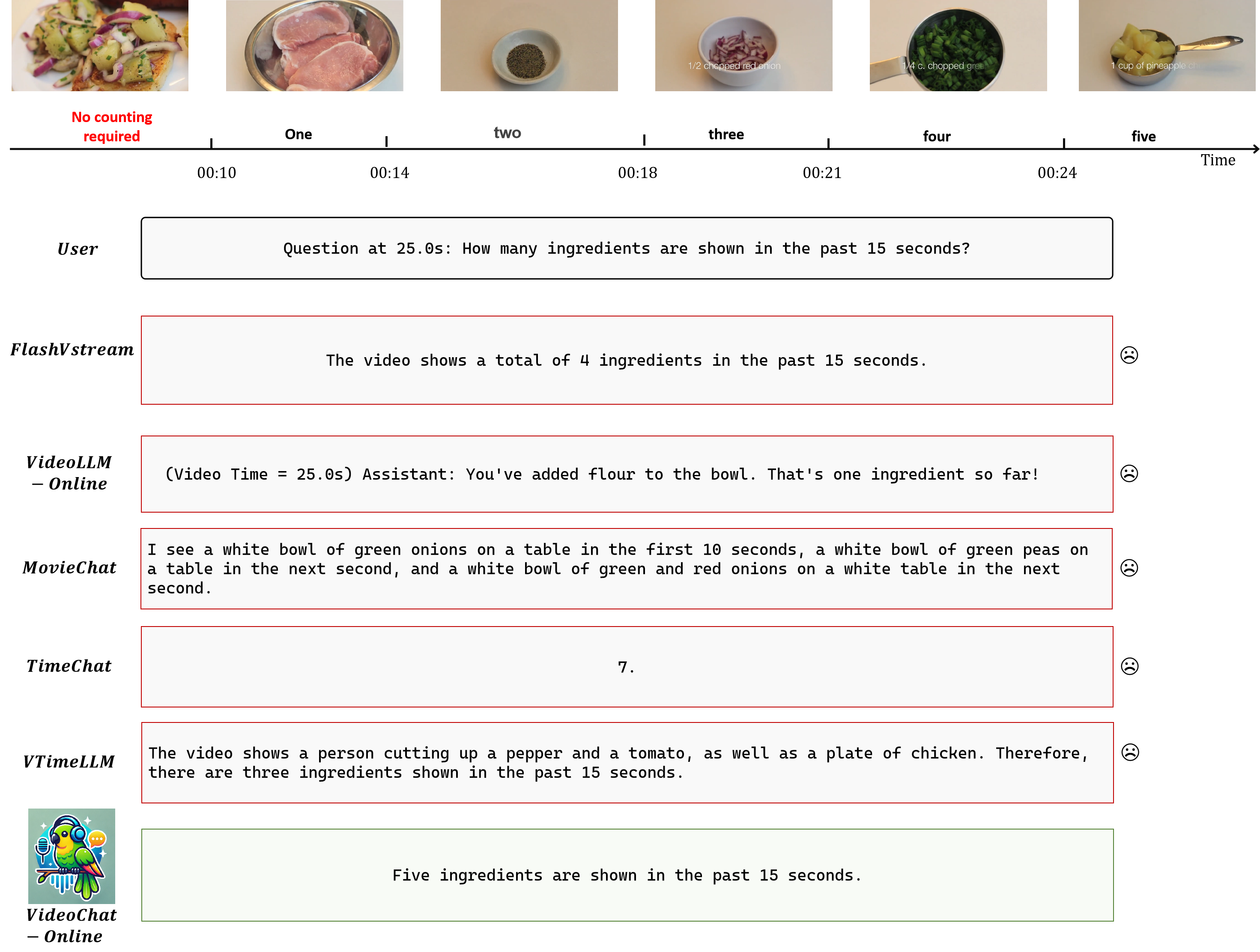}
    \caption{Qualitative Results}
    \label{fig:case}
\end{figure*}

\textbf{Qualitative comparison.} We provide a qualitative comparison with other online models in Figure~\ref{fig:case}. Including TimeChat~\cite{timechat} and VTimeLLM~\cite{huang2024vtimellm}, which are time-sensitive models, and Flash-VStream\cite{flashvstream}, VideoLLM-Online\cite{videollm_online} and MovieChat\cite{moviechat}, which can receive streaming input.

\section{More Ablations}

\subsection{Hierarchical Memory Bank Visualization}
Figure~\ref{fig:memory_bank} provides a visualization example of the proposed hierarchical memory bank, where frames in the $m_s$ layer are marked in red, those in the main layer are marked in orange, and the others belong to the $m_t$ layer. The corresponding capacity of the memory bank, $\left[m_t, m_{\text{main}}, m_s\right]$, is $\left[12, 2, 2\right]$.

\subsection{Online SFT Data}
Table~\ref{tab:ablation-data} shows the ablation experimental results of the online instruction fine-tuning dataset under different task categories. For simplicity, in the experiment, only 1 epoch of instruction fine-tuning was performed using the online dataset. As can be seen from the results, the complete online dataset (Online Data SFT) achieved the highest performance on \benchname (48.1).
The performance of each task module decreased after removal, with the removal of w/o dense captioning and w/o step localization decreasing by 1.1 and 1.4 percentage points, respectively. However, the removal of w/o spatial-temporal action detection and w/o temporal grounding resulted in more significant performance degradation, with a decrease of 3.3 and 2.7 percentage points, respectively.

\subsection{Model}
\textbf{Memory capacity.} As shown in the Table \ref{tab:supp_mem_cap}, we expand the memory capacity to {3, 3, 18} and {4, 4, 24}, and the existing memory capacity settings maintain optimal performance and computational efficiency.
\subsection{Benchmark}

\noindent\textbf{The impact of fps.}
We fixed the sampling frame number to 64, selected fps as \{0.25, 0.5, 1, 2, 4\}, and uniform sampling (from the beginning of the video to the question timestamp). The impact of fps on model performance under the sliding window setting is shown in Figure~\ref{fig:vary-fps}. \textbf{Higher fps offers better performance.}

\noindent\textbf{The impact of input frames (sliding window size).}
We fixed the fps to 2 and selected 16, 32, and 64 frames for evaluation in Figure~\ref{fig:vary-frame}. We select LongVA~\cite{longva}, trained exclusively on static image data, LLaMA-VID~\cite{llamavid}, which incorporates both single-image and video training data, and MLLM, an extension of LLaVA-OneVision~\cite{LLaVA-OneVision} trained on single-image, multi-image, and video data, for a comprehensive comparison. Notably, the advantages of our model in handling diverse task types and achieving superior overall performance remain consistent regardless of the number of frames. This demonstrates the value of online data in enhancing performance in real-time scenarios, while minimizing computational overhead, which expands deployment possibilities.

\noindent\textbf{The impact of model size.}
We use InternVL2~\cite{chen2023internvl} family as the research object as it has a wide variety of models of different scales: \{1B, 2B, 4B, 8B\}.

As shown in Table \ref{tab:ovbench_scale}, it can be seen that the performance of models 1, 2, and 4B is almost the same, but there has been significant improvement in performance for the 8B model. It is crucial to deploy larger-scale models in online scenarios effectively.

\section{Benchmark Details}

\subsection{Video and Query Length Distributions}  
\begin{figure*}[ht]
    \centering
    \begin{minipage}{0.3\linewidth}
        \setlength{\intextsep}{0pt} 
        \setlength{\abovecaptionskip}{0pt}
        \centering
        \includegraphics[width=1.1\linewidth]{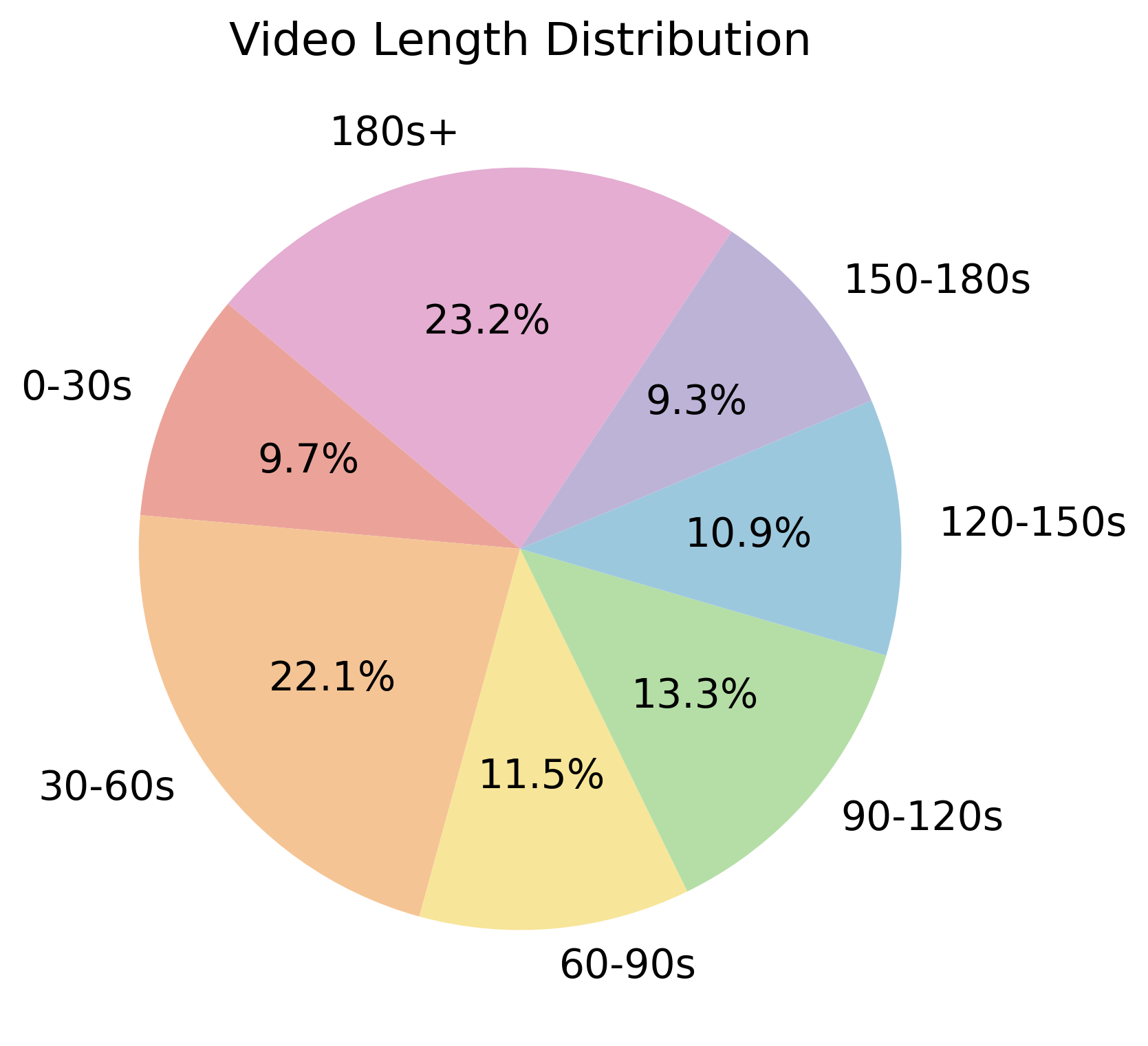}
        \label{fig:video}
    \end{minipage}%
    \hfill
    \begin{minipage}{0.5\linewidth}
        \centering
        \includegraphics[width=1.\linewidth]{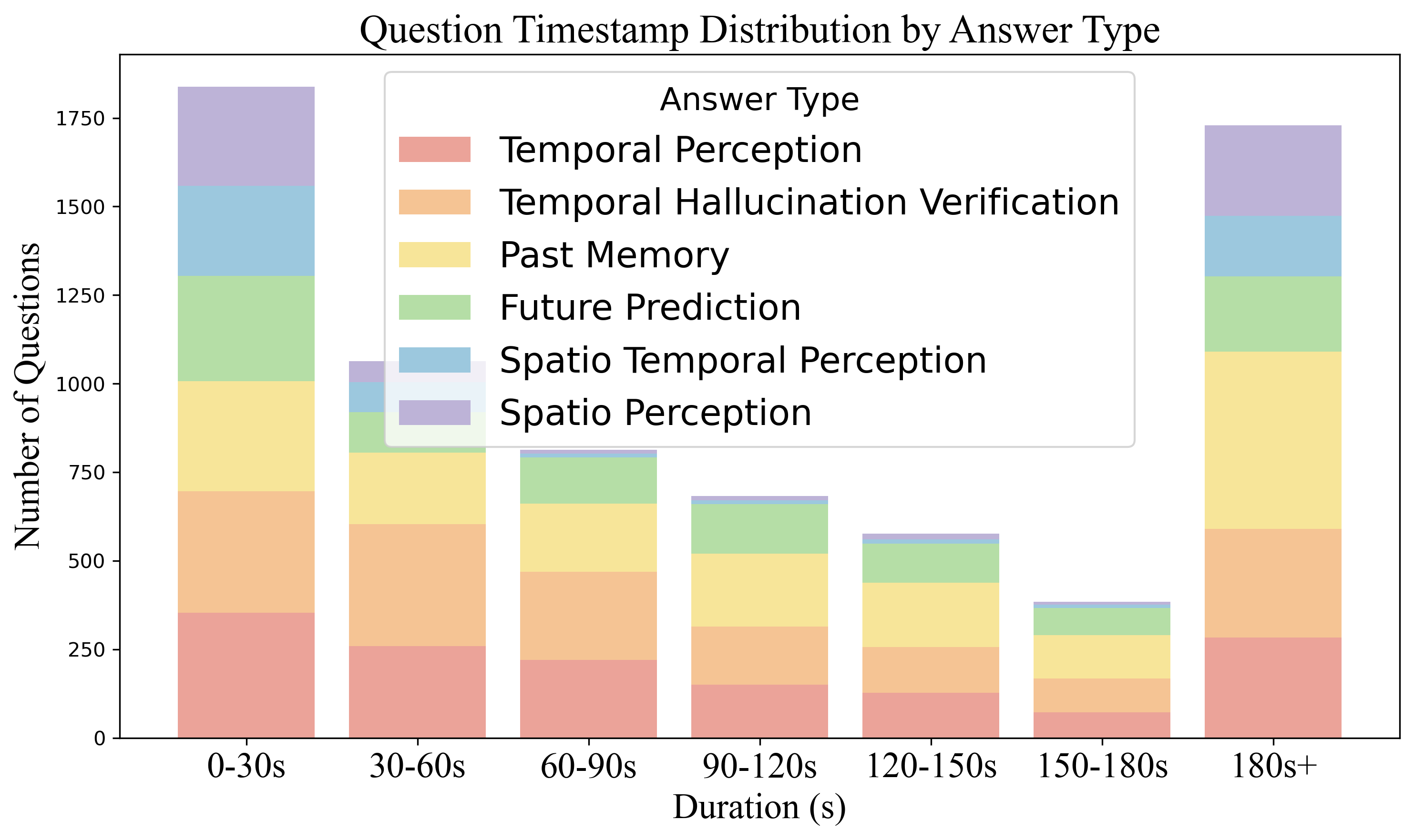}
        \label{fig:question}
    \end{minipage}
    \caption{Distributions of video and query lengths. The left figure represents the video length distribution, while the right figure shows the query length distribution.}
    \label{fig:video_query_dist}
\end{figure*}

Total 1,463 videos. The distributions of video lengths and query lengths are illustrated in Figure~\ref{fig:video_query_dist}. 

\subsection{Details of QA Generation}
The QA template for OVBench is shown in Table~\ref{tab:task_templates}. For each task type with different detailed spatiotemporal annotations, we have taken specific measures in Table~\ref{tab:qa_generation} to ensure the diversity and difficulty of the problem and option generation.

\subsection{Data Examples}

\begin{figure*}
    \centering
    \includegraphics[width=0.8\textwidth]{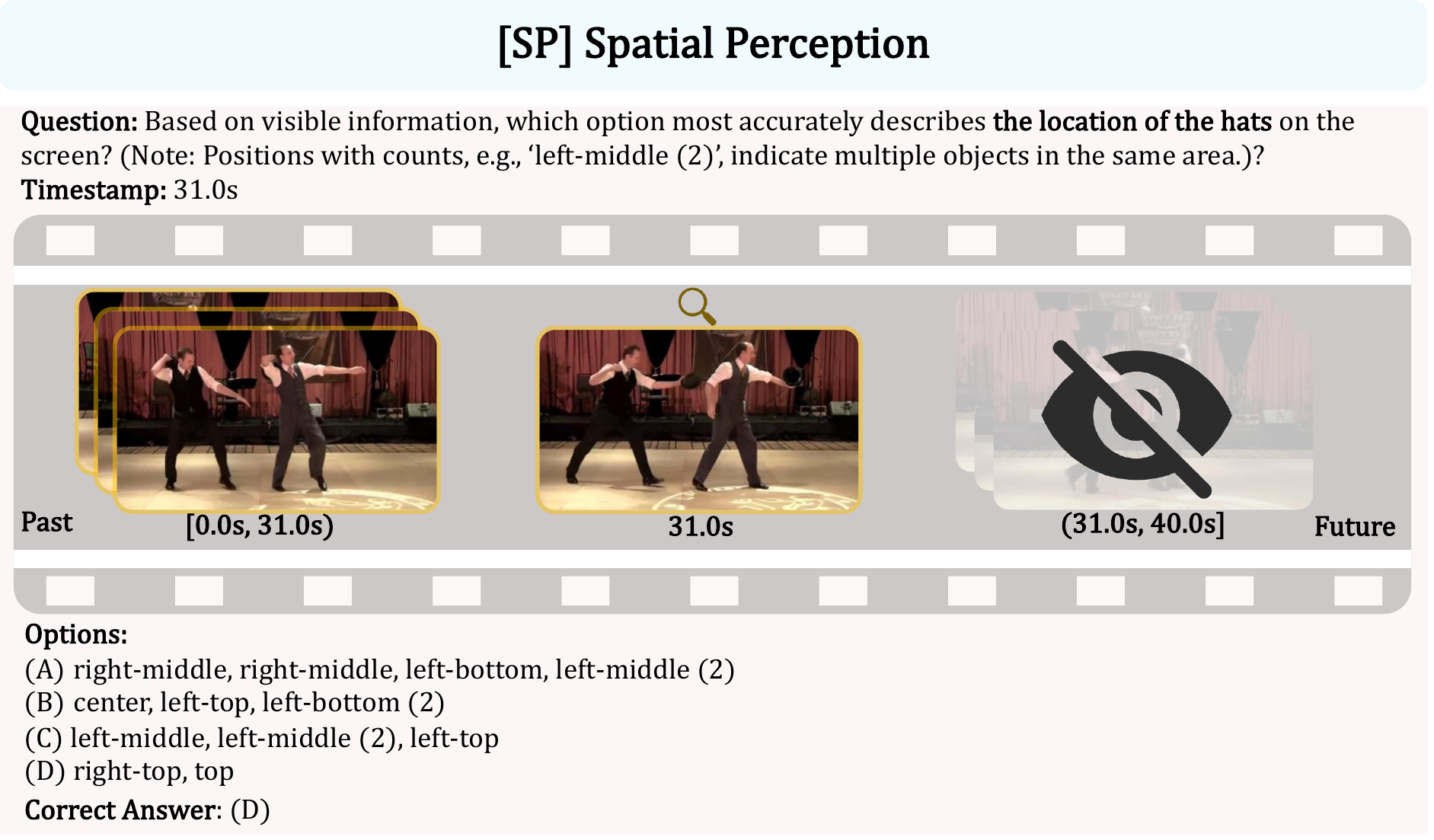}
    \vspace{0.5em} 
    \includegraphics[width=0.8\textwidth]{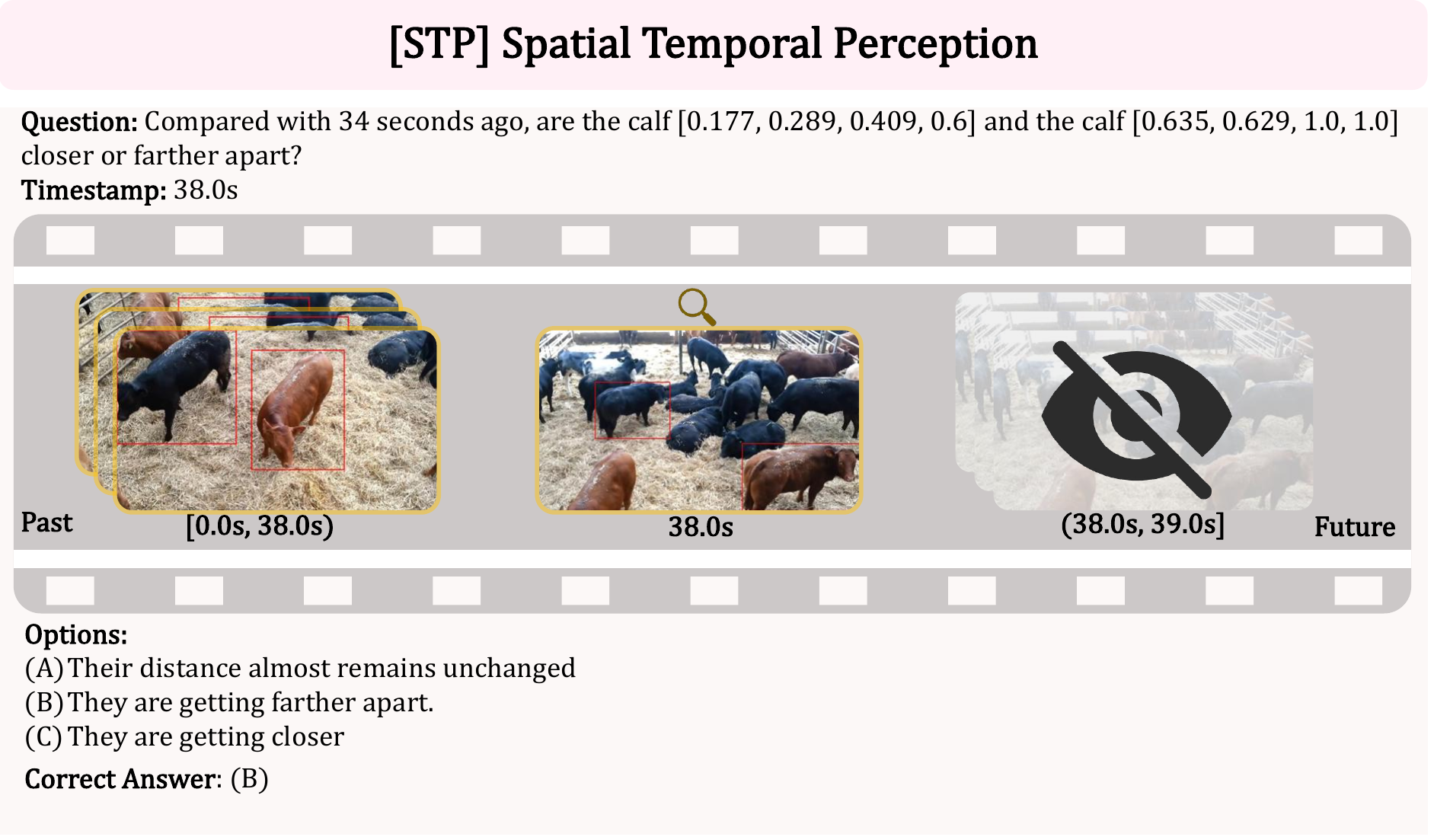}
    \vspace{0.5em}
    \caption{Task examples in \bench: Spatial Perception\&Spatial Temporal Perception}
    \label{fig:combined-benchmarks-1}
\end{figure*}

\begin{figure*}
    \centering
    \includegraphics[width=0.8\textwidth]{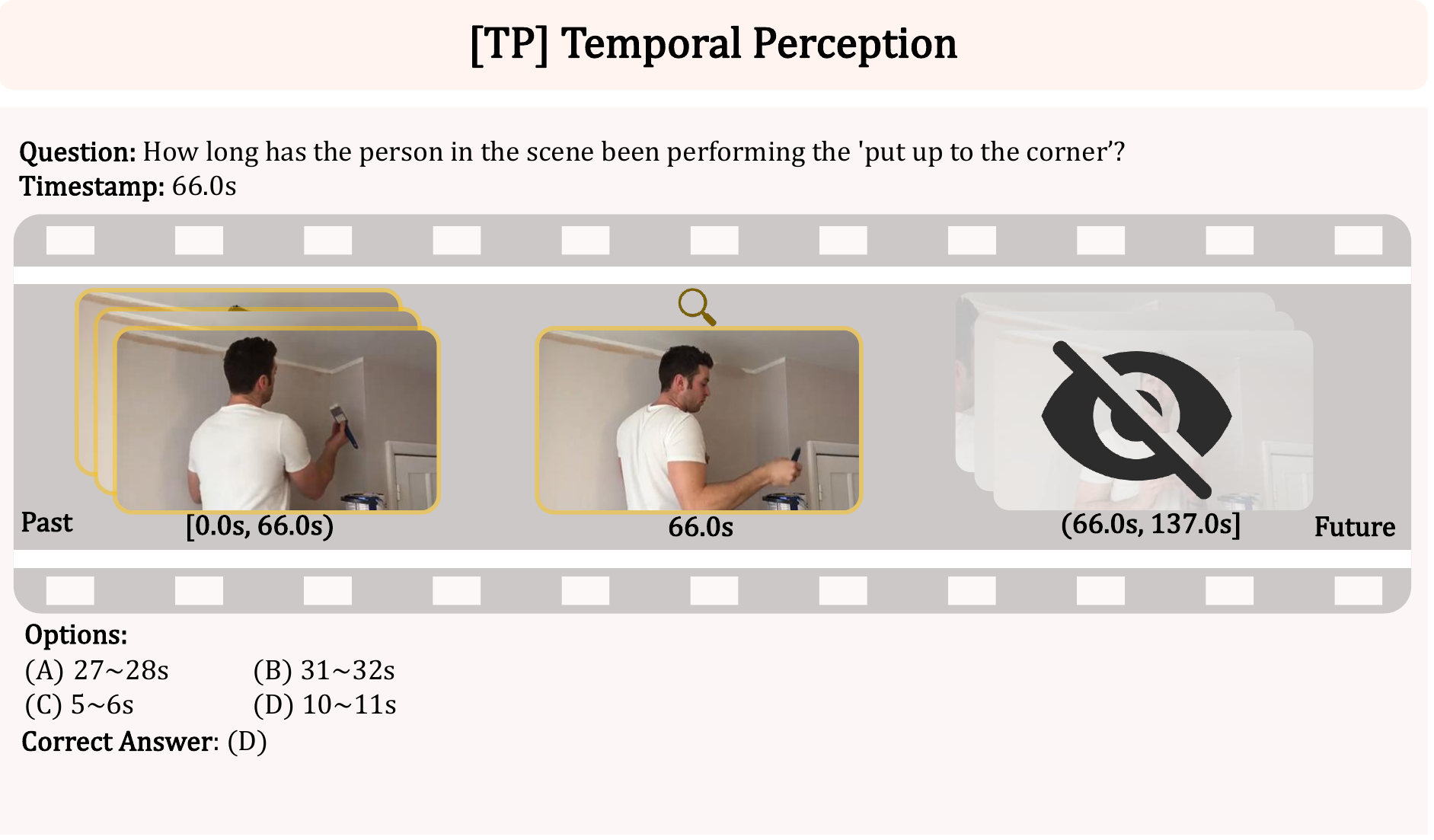}
    \vspace{0.5em}
    \includegraphics[width=0.8\textwidth]{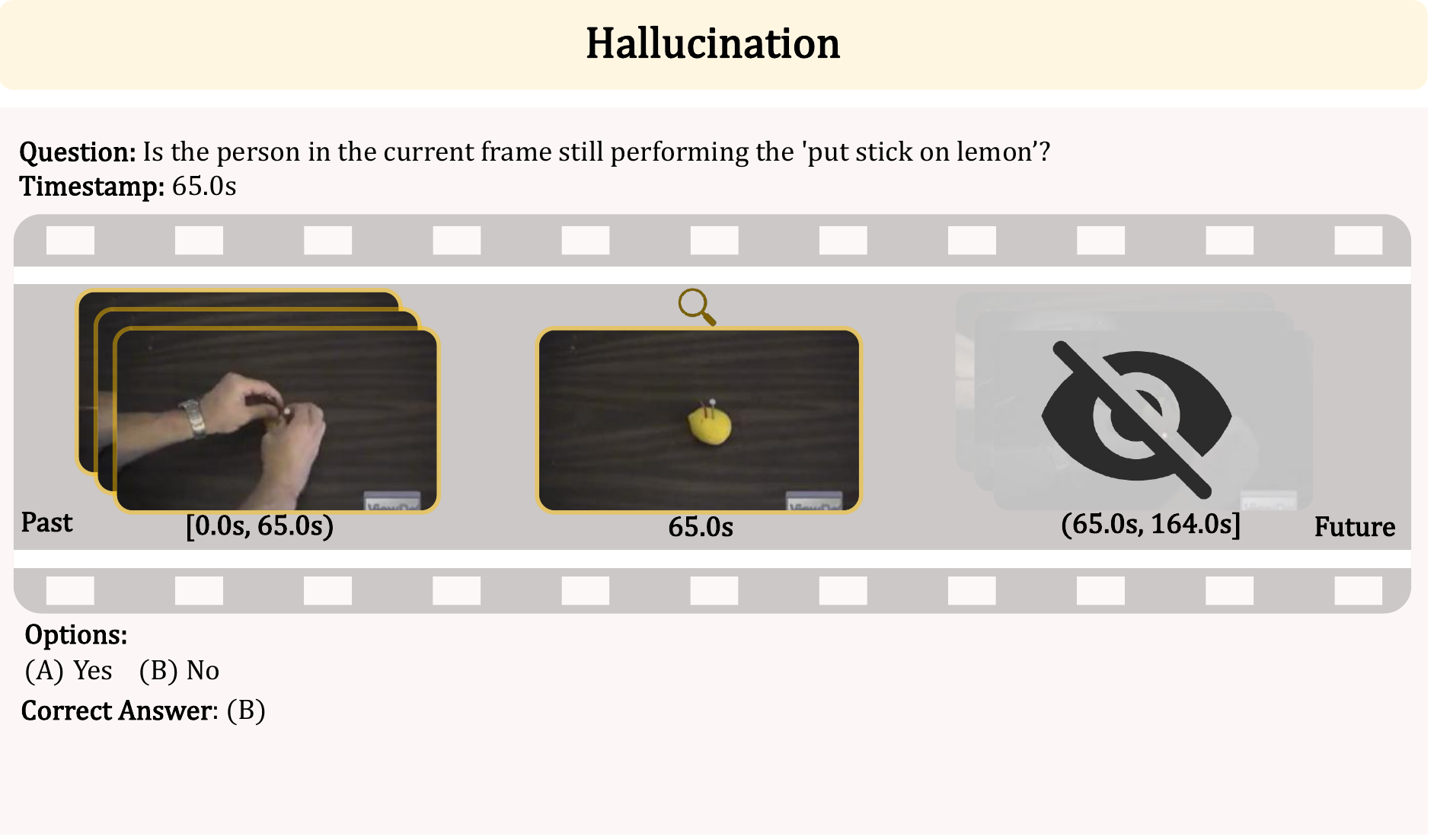}
    \vspace{0.5em}
    \caption{Task examples in \bench: Temporal Perception\&Temporal Hallucination Verification}
    \label{fig:combined-benchmarks-2}
\end{figure*}

\begin{figure*}
    \centering
    \includegraphics[width=0.8\textwidth]{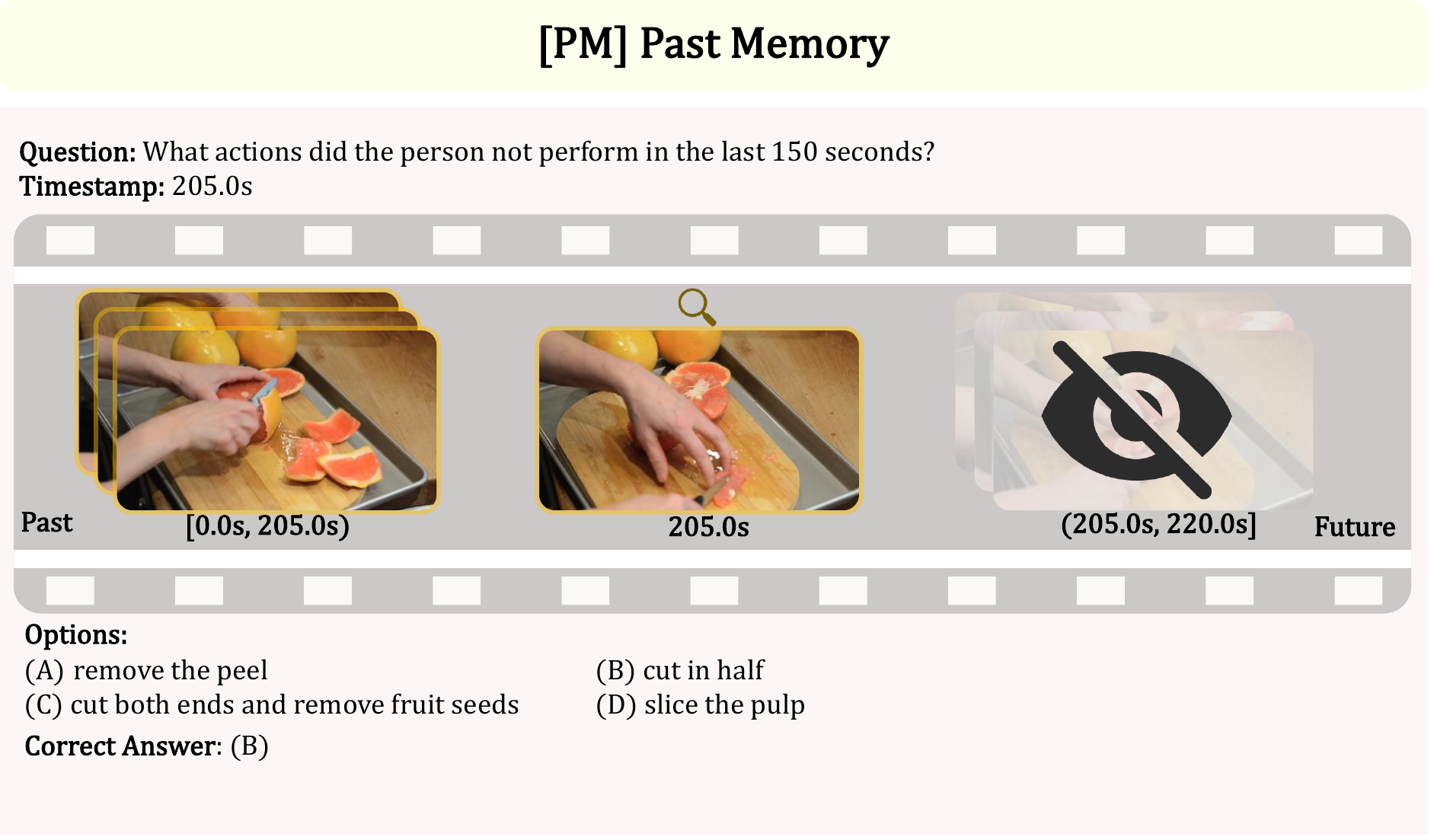}
    \vspace{0.5em}
    \includegraphics[width=0.8\textwidth]{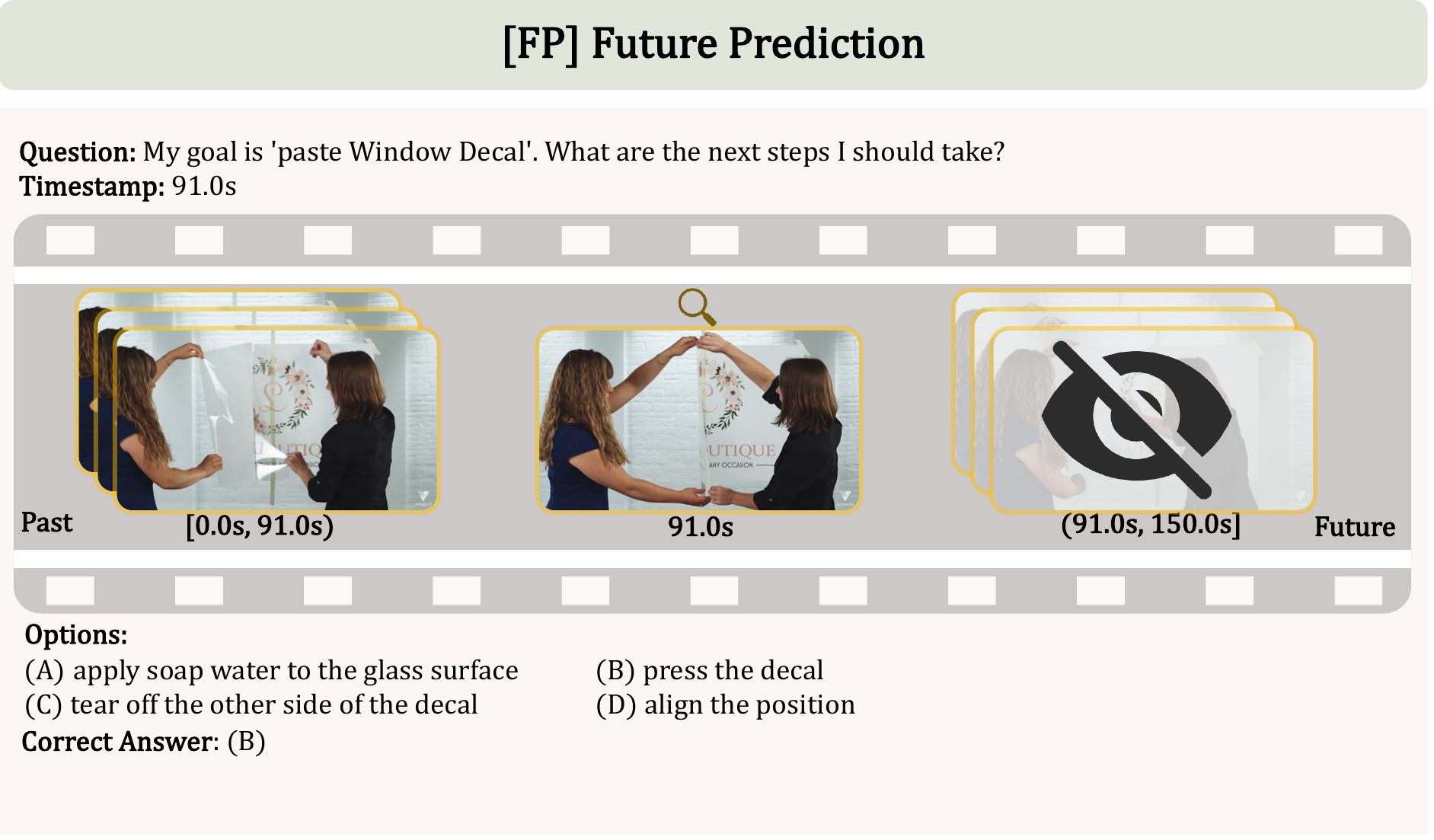}
    \caption{Task examples in \bench: Past Memory\&Future Prediction}
    \label{fig:combined-benchmarks-3}
\end{figure*}

One visual example for each task type, as shown in Figure \ref{fig:combined-benchmarks-1}, \ref{fig:combined-benchmarks-2}, and \ref{fig:combined-benchmarks-3}.

\section{Training and Inference Hyper-parameters}

The hyperparameters used in training and the memory bank fps and capacity settings during inference are shown in the table~\ref{tab:train_hype} and table ~\ref{tab:infer_hype}.



\end{document}